\documentclass[iicol,sn-basic,Numbered]{sn-jnl}

\usepackage{graphicx}%
\usepackage{multirow}%
\usepackage{amsmath,amssymb,amsfonts}%
\usepackage{amsthm}%
\usepackage{mathrsfs}%
\usepackage[title]{appendix}%
\usepackage{xcolor}%
\usepackage{textcomp}%
\usepackage{manyfoot}%
\usepackage{booktabs}%
\usepackage{algorithm}%
\usepackage{algorithmicx}%
\usepackage{algpseudocode}%
\usepackage{listings}%

\theoremstyle{thmstyleone}%
\theoremstyle{thmstyletwo}%

\theoremstyle{thmstylethree}%

\def\eg{\emph{e.g.}} 
\def\ie{\emph{i.e.}}

\title{InCTRLv2: Generalist Residual Models for Few-Shot Anomaly Detection and Segmentation}

\author[1]{\fnm{Jiawen} \sur{Zhu}}\email{jwzhu.2022@phdcs.smu.edu.sg}
\author[2]{\fnm{Mengjia} \sur{Niu}}\email{m.niu21@imperial.ac.uk}
\author*[1]{\fnm{Guansong} \sur{Pang}}\email{gspang@smu.edu.sg}

\affil*[1]{\orgdiv{School of Computing and Information Systems}, \orgname{Singapore Management University}, \orgaddress{\country{Singapore}}}
\affil[2]{\orgdiv{Dyson School of Design Engineering}, \orgname{Imperial College London}, \orgaddress{\country{United Kingdom}}}

\abstract{
While recent anomaly detection (AD) methods have made substantial progress in recognizing abnormal patterns within specific domains, most of them are \textit{specialist models} that are trained on large training samples from a specific target dataset, struggling to generalize to unseen datasets. To address this limitation, the paradigm of Generalist Anomaly Detection (GAD) has emerged in recent years, aiming to learn a single \textit{generalist model} to detect anomalies across diverse domains without retraining.
To this end, this work introduces InCTRLv2, a novel few-shot Generalist Anomaly Detection and Segmentation (GADS) framework that significantly extends our previously proposed GAD model, InCTRL.
Building on the idea of learning in-context residuals with few-shot normal examples to detect anomalies as in InCTRL, InCTRLv2 introduces two new, complementary perspectives of anomaly perception under a dual-branch framework.
This is accomplished by two novel modules upon InCTRL: 
\textit{i) Discriminative Anomaly Score Learning (DASL) with both normal and abnormal data in the main branch}, which learns a semantic-guided abnormality and normality space that supports the classification of query samples from both the abnormality and normality perspectives;
and \textit{ii) One-class Anomaly Score Learning (OASL) using only the normal data}, 
which learns generalized normality patterns in a semantic space via an auxiliary branch, 
focusing on detecting anomalies through the lens of normality solely.  
Both branches are guided by rich visual–text semantic priors encoded by large-scale vision–language models, enabling robust semantic grounding beyond domain-specific visual appearances. Together, these two branches offer a dual semantic perspective for AD: one emphasizes normal-abnormal discriminative power, while the other emphasizes semantic of being deviated from the normality.
Extensive experiments on ten real-world AD datasets, covering industrial defects, medical anomalies, and semantic anomalies, demonstrate that InCTRLv2 achieves state-of-the-art performance in both anomaly detection and segmentation tasks across one-, two-, and four-shot settings. Code is available at \renewcommand\UrlFont{\color{blue}}\url{https://github.com/mala-lab/InCTRLv2}.
}

\begin{document}

\maketitle

\section{Introduction}
Anomaly Detection (AD) aims to detect data that significantly deviate from the majority of samples in a dataset. It is a vital task in computer vision and showcases a wide range of real-world applications such as industrial inspection, medical imaging analysis, and scientific discovery \cite{pang2021deep, cao2024survey}. Owing to this significance, numerous AD approaches have been introduced over the years, such as reconstruction-based models \cite{akcay2019ganomaly, schlegl2019f, zavrtanik2021reconstruction, yan2021learning, zaheer2020old, zavrtanik2021draem, park2020learning, hou2021divide, xiang2023squid, liu2023diversity, yao2023one, yao2023focus} and one-class classification techniques \cite{tax2004support, yi2020patch, bergman2020classification, chen2022deep, ruff2020deep, Ye_2025_CVPR}, but they are mostly specialist methods, relying on dataset-specific training with large normal data for each AD task in a target domain, \ie, one AD model for each dataset. Such approaches operate under a closed-set distribution assumption---training and test data are from the same distribution---and often exhibit poor generalization to new, unseen datasets, primarily due to inherent dataset-specific biases learned in these models. 
However, in real-world scenarios, collecting large-scale labeled normal data for every application scenario is often impractical, particularly in sensitive domains such as healthcare or manufacturing, where data privacy and operational constraints exist. This limitation highlights the need for more generalizable anomaly detection approaches.

Recent advances in large Vision–Language Models (VLMs) have demonstrated strong
generalization capabilities owing to their pretraining on web-scale image–text datasets. 
Early attempts to leverage VLMs for anomaly detection include WinCLIP~\cite{jeong2023winclip}, which is in a training-free manner to improve transferability in zero- and few-shot settings. 
While effective, WinCLIP relies heavily on VLM’s (\textit{i.e.}, CLIP~\cite{radford2021learning}) pretrained generalization and handcrafted prompts, which are often tuned to domain-specific defects (\textit{e.g.}, industrial anomalies), thereby limiting its applicability across diverse domains such as medical imaging or semantic anomaly detection. AprilGAN~\cite{chen2023april} then introduces learnable parameters to better adapt CLIP to downstream AD tasks, yet it remains similarly constrained by reliance on prompt engineering and domain bias.

Subsequently, PromptAD~\cite{li2024promptad} and our earlier work InCTRL~\cite{zhu2024generalistanomalydetectionincontext} advance this line of research in the few-shot anomaly detection (FSAD) setting, but in different directions. PromptAD investigates the FSAD setting by incorporating learnable prompts, but it still adheres to the specialist model paradigm, where a separate model is required for each target dataset. In contrast, InCTRL pioneers the paradigm of \textbf{Generalist Anomaly Detection} (GAD), aiming to \textit{train one single model that can generalize to detect anomalies in diverse datasets from different application domains without any further training on the target data}. 

To achieve generalization capability across domains, an in-context residual learning framework is introduced in InCTRL~\cite{zhu2024generalistanomalydetectionincontext} , in which a query sample is compared against a small set of in-context normal sample prompts (\textit{i.e.}, normal reference samples) to model domain-agnostic residuals, with the in-context visual prompts serving as representations of normality. The core idea of InCTRL is to model the discrepancy between a query sample and these normal references. As a result, abnormal samples yield larger residuals, enabling transferable detection across diverse domains. Moreover, InCTRL models residuals at both the image and patch levels, capturing coarse- and fine-grained anomaly cues under image-level supervision to improve generalization. More recently, ResAD~\cite{yao2024resad} also exploits residual learning for GAD, but it adopts a fundamentally different formulation of residuals.  

Despite its strengths, InCTRL faces a key limitation exposed by cross-domain deployment: the residual learning operates purely in the visual feature space and fails to capture semantic knowledge about abnormality beyond those reference-based residuals, making it sensitive to superficial appearance changes (\textit{e.g.}, differences in texture, illumination, or material). Such domain-specific variations can be mistaken for anomalies, limiting generalization to unseen datasets. In addition, the nature of abnormality varies substantially across domains, such as a stain in wood versus a crack in a capsule, so a single discriminative boundary is potentially unstable. In contrast, research in the one-class anomaly detection area has demonstrated that the definition of normality tends to exhibit consistent statistical properties across domains. However, InCTRL fails to explicitly model such a domain-stable normality manifold, further limiting its robustness.

\begin{figure}[t]
    \centering
    \includegraphics[width=0.45\textwidth]{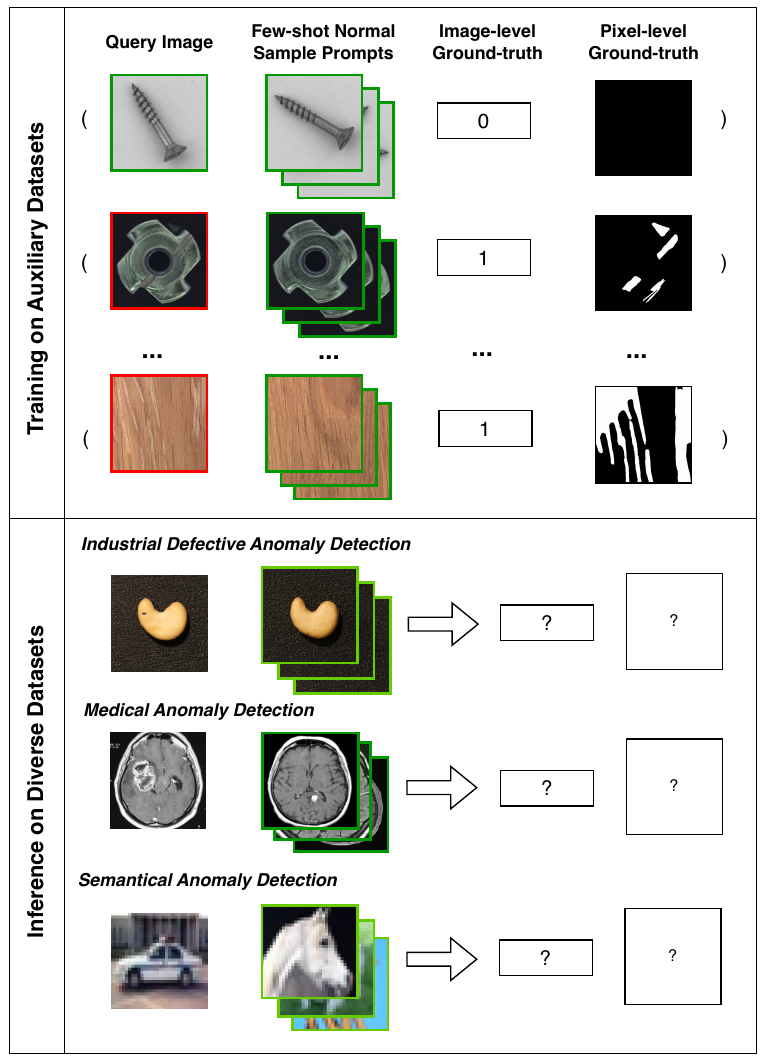}
    \caption{Illustration of the Generalist Anomaly Detection and Segmentation (GADS) paradigm for the few-shot AD task. A model is trained on auxiliary datasets and leverages few-shot normal images as in-context sample prompts paired with query samples (\textbf{Top}). It can then be directly applied to diverse target datasets from different domains without requiring domain-specific retraining (\textbf{Bottom}).}
    \label{fig:intro}
\end{figure}
To address these issues, we propose InCTRLv2, a significantly enhanced extension of InCTRL that augments the original framework with a semantic-guided joint learning strategy built on CLIP text priors, thereby equipping generalist anomaly detection with complementary semantic cues for anomaly understanding. In addition, we generalize the original formulation from GAD to generalist anomaly detection and segmentation (GADS), allowing the model to capture both image-level and pixel-level anomalies in a unified manner, as illustrated in Fig. ~\ref{fig:intro}. To achieve this, InCTRLv2 incorporates in-context residual learning with two complementary semantic-guided branches. \textbf{i)} The first one is a discriminative anomaly score learning (DASL) module with both abnormality and normality data. This module aligns visual features with hand-crafted normal and abnormal text prototypes to compute semantic-guided anomaly scores and maps for each query image. Through this alignment, DASL constructs a new discrepancy decision space between normality and abnormality with semantic priors, providing complementary signals of abnormality to what is extracted by in-context residual learning. \textbf{ii)} Another branch is instantiated by a one-class anomaly score learning (OASL) module with only normality data. 
Inspired by the aforementioned observation that the definition of abnormality often varies across domains while normality presents a stable tendency in previous one-class AD approaches \cite{ruff2018deep}, OASL utilizes normal data only to distill a compact one-class decision space that captures the intrinsic characteristics of normality and obtain associated one-class abnormality maps for anomaly scoring, enabling more stable cross-domain AD performance.

By combining these modules, InCTRLv2 unifies three complementary signals: \textbf{i)} residual deviations from few-shot normal prototypes, \textbf{ii)} semantic-guided discriminative anomaly understanding grounded in discrepancies between normality and abnormality, and \textbf{iii)} semantic-guided one-class anomaly understanding distilled from intrinsic distribution of normality. This unified formulation not only enriches the discriminative space between normal and abnormal samples with semantic priors but also enhances cross-domain generalization through an invariant normality perspective. Consequently, InCTRLv2 achieves accurate and robust anomaly detection and segmentation under various few-shot settings.

Our main contributions are summarized as follows:
\begin{itemize}

\item We extend the original InCTRL framework to InCTRLv2, which couples fine-grained in-context residual learning with a semantic-guided joint learning framework, involving a discriminative anomaly score learning (DASL) module with both abnormality and normality data and a one-class anomaly score learning (OASL) module with only normality data. This enriches the residual-based feature space and enhances cross-domain generalization.

\item We propose the DASL module, a semantic-guided branch to align the visual representations with text prompts for modeling both normality and abnormality, thereby providing complementary anomaly cues for residual learning and establishing a semantically discriminative decision space between normal and abnormal samples.

\item We introduce the OASL module, a semantic-guided one-class learning branch trained exclusively on normal samples to derive a compact decision space that encapsulates the intrinsic characteristics of normality.
By exploiting the inherent stability and domain invariance of normal patterns, 
OASL emphasizes
semantic regularization grounded in the intrinsic distribution of normality.

\item We conduct extensive experiments on ten benchmark datasets under one-, two-, and four-shot settings. Results show that InCTRLv2 achieves state-of-the-art performance in both anomaly detection and segmentation, while substantially improving the performance of the original InCTRL.
\end{itemize}

\section{Related Works}
\subsection{Anomaly Detection}
Traditional anomaly detection (AD) approaches are predominantly designed under the assumption that models are trained and deployed on the same domain, with access to sufficient normal training data~\cite{pang2021deep,wu2024deep,cao2024survey,zhu2024anomaly}. These methods can be broadly categorized into several types. One-class classification methods~\cite{tax2004support, yi2020patch, bergman2020classification, chen2022deep, ruff2020deep} aim to tightly enclose normal data distributions using techniques like support vector, such that samples deviating from this boundary can be flagged as anomalies. Reconstruction-based methods~\cite{akcay2019ganomaly, schlegl2019f, zavrtanik2021reconstruction, yan2021learning, zaheer2020old, zavrtanik2021draem, park2020learning, hou2021divide, xiang2023squid, liu2023diversity, yao2023one, yao2023focus} attempt to learn a generative model that reconstructs normal samples accurately, then the anomalies can be detected based on their higher reconstruction errors.
Distance-based approaches~\cite{pang2018learning,defard2021padim, cohen2020sub, roth2022towards} detect anomalies by computing distances between features of test samples and stored normal training features using pre-trained encoders. These methods leverage the inductive bias of pretrained models but still require a sufficiently large target domain normal dataset to build reliable feature memory bank. Knowledge distillation methods~\cite{deng2022anomaly, bergmann2020uninformed, salehi2021multiresolution, wang2021student, Cao_2023_ICCV, tien2023revisiting, zhang2023destseg} distill knowledge from a fixed teacher model to a student model trained only on normal data. Discrepancies between the teacher and student outputs are then used to detect anomalies.

While these techniques have demonstrated success in controlled settings, they typically operate in a one-model-per-dataset regime, limiting their applicability in scenarios with domain shifts or unknown anomalies. Recent works have begun exploring anomaly detection under domain shift or few-shot adaptation~\cite{li2024promptad}, but they often assume high relevance between the source and target domains or require access to domain-specific data during training. In contrast to these prior paradigms, our goal is generalist anomaly detection and segmentation (GADS), where a single model can generalize to detect diverse anomalies across unseen domains without training or adaptation on target data. There are a few concurrent studies leverage vision-language models (VLMs) for AD, some of they operate under different assumptions, such as weak supervision~\cite{wu2023open,wu2023vadclip} or direct zero-shot transfer without in-context visual adaptation~\cite{zhou2023anomalyclip, gu2024filo, zhu2024fine}.

\subsection{Few-shot Anomaly Detection}
Few-shot anomaly detection (FSAD) aims to detect anomalies with access to only a few set of normal samples from the target domain. Traditional FSAD methods~\cite{sheynin2021hierarchical, huang2022registration, wu2021learning, belton2023fewsome, schwartz2022maeday, wang2022few, xie2023pushing,liao2024coft} attempt to model the distribution of these few samples using reconstruction or embedding-based strategies, often requiring re-training or fine-tuning to fit in new domains, which limits their scalability and generalization. Distance-based approaches such as SPADE~\cite{cohen2020sub}, PaDiM~\cite{defard2021padim}, and PatchCore~\cite{roth2022towards} alleviate this by leveraging pre-trained image features and computing distances between query features and those from few-shot support images. While effective, these methods still assume the anomaly distribution lies close to the representation space of the pretrained encoder, where the learned knowledge may not transfer reliably across domains with distinct characteristics.

RegAD~\cite{huang2022registration} offers a domain-agnostic few-shot detection mechanism using registration-based feature matching, but its performance depends on domain similarity between training and test sets. WinCLIP~\cite{jeong2023winclip} represents an important step toward using large vision-language models for FSAD. It employs handcrafted text prompts and multi-scale visual features to detect anomalies with CLIP~\cite{li2023clip}. While WinCLIP achieves promising results on industrial defect datasets, it struggles in generalization, particularly in settings where handcrafted prompts do not align well with the semantic characteristics of anomalies—such as in medical or semantic domains. 

Following this line of research, several FSAD methods, such as AprilGAN~\cite{chen2023april} and One-for-Normal~\cite{li2024one} have further explored few-shot anomaly detection with the aid of VLMs. However, most of these works overlook the modeling of residual information between normal and anomalous samples. PromptAD~\cite{li2024promptad} proposes a one-class prompt learning method, but its FSAD setting requires training on few-shot samples. Our earlier work InCTRL~\cite{zhu2024generalistanomalydetectionincontext} introduced a novel in-context residual learning framework that models the generalizable discrepancies between query samples (normal or abnormal) and a few-shot set of normal patterns. This design allows the model to capture transferable residual signals without requiring domain-specific retraining.
ResAD~\cite{yao2024resad} also incorporates residual information to address the same FSAD setting as InCTRL, but it was proposed later than our work and defines residuals in a fundamentally different manner.

\subsection{In-context Learning}
In-context learning (ICL) has shown remarkable success in NLP by enabling large language models to adapt to novel tasks using a few in-context examples~\cite{alayrac2022flamingo, brown2020language, hao2022language}. Extending this idea to vision, recent works~\cite{chen2021pix2seq, chen2022unified, kolesnikov2022uvim, lu2022unified, wang2022ofa} reformulate vision tasks into language problems using tokenized prompts or image-to-text mapping. Previous works like Visual Prompting~\cite{bar2022visual} and Painter~\cite{wang2023images, wang2023seggpt} explore grid-based masked inpainting or visual prompting for dense prediction tasks. However, these methods mainly focus on task-level generalization (i.e., adapting to new segmentation or classification tasks) and are not directly suitable for instance-level anomaly detection, where fine-grained discrepancy modeling is critical. 

Our work departs from this line by redesigning in-context learning specifically for GADS. Instead of prompting the model with tasks, we treat few-shot normal images as visual in-context prompts that define dataset-specific normality. We then measure residuals between query and in-context features, capturing visual discrepancies that indicate anomalous content. This formulation enables our model to generalize across different datasets by learning from a unified framework of in-context residuals, rather than retraining or relying on fixed prompt templates.

\begin{figure*}[ht]
    \centering
    \includegraphics[width=\textwidth]{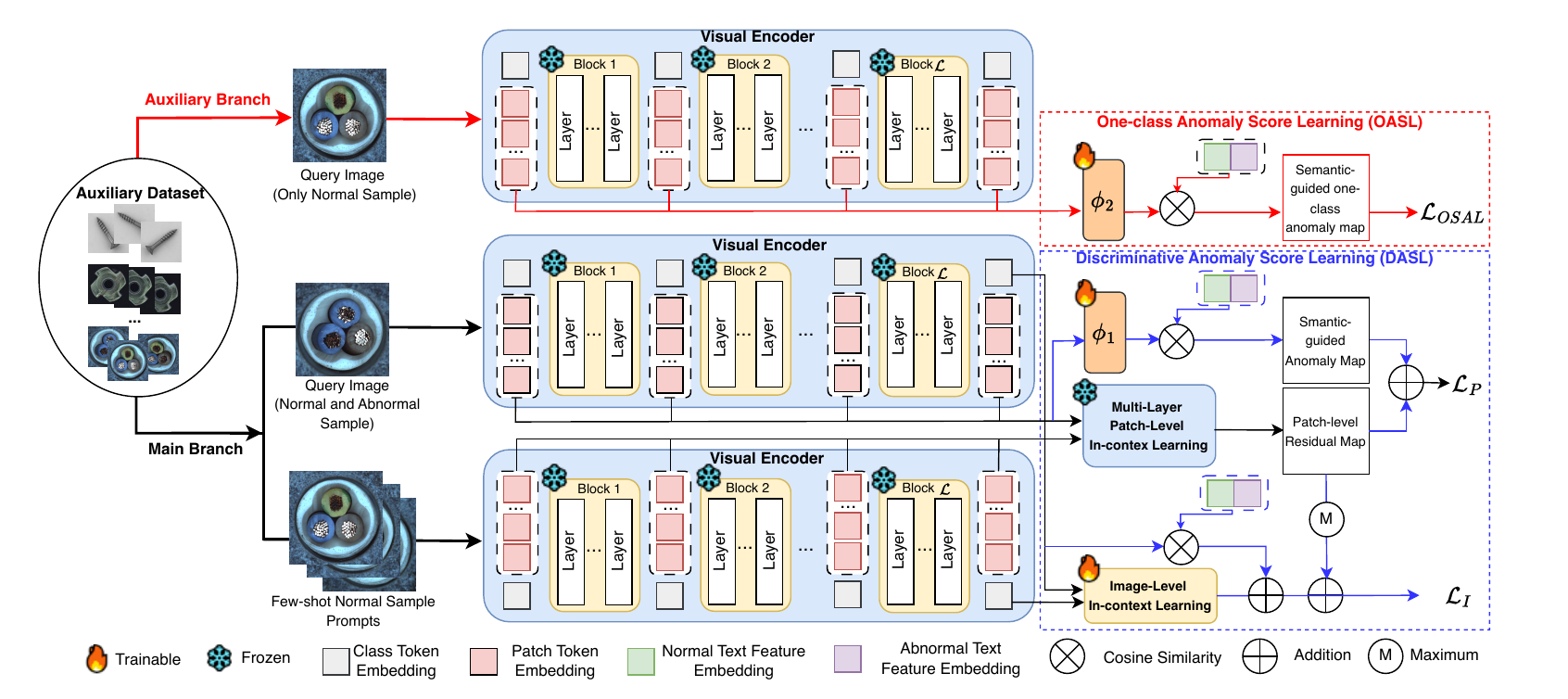}
    \caption{Overview of the training process of InCTRLv2. It extends the single-branch residual learning framework of InCTRL into a dual-branch architecture, consisting of a main branch and an auxiliary branch. The main branch employs a Discriminative Anomaly Score Learning (DASL) module to learn a semantic-guided decision space that jointly models abnormality and normality, enabling anomaly discrimination from both perspectives. In parallel, the auxiliary branch adopts a One-class Anomaly Score Learning (OASL) module, which is trained exclusively on normal samples to capture generalized normality patterns in the semantic space. Together, these two branches complement each other by combining discriminative abnormality modeling with normality-driven guidance.
    }
    \label{fig:overall}
\end{figure*}

\section{Preliminaries}
\subsection{Problem Statement.}
Generalist Anomaly Detection and Segmentation (GADS) aims to train a single model that can effectively detect and segment anomalies on test datasets from diverse application domains without requiring any training on the target data. To this end, we treat the training data as an auxiliary dataset, which is assumed to be drawn from distributions that are different from those of the target test sets. Formally, let $\mathcal{D}_{train} = \{X_{train}, Y_{train}\}$ denote an \textit{auxiliary} training dataset, where 
$ X_{train} = \{x_i\}_{i=1}^N, x_i\in \mathbb{R}^{h \times w \times 3} $
is a set of $N$ training images, with $h$ and $w$ respectively representing the height and width of image sample and $ Y_{train} = \{ y_i, G_i\}_{i=1}^N $ represents the corresponding image-level labels and pixel-level anomaly masks. Each image $x_i$ is associated with a binary label $y_i \in \{0,1\}$, indicating whether the image is normal ($y_i=0$) or abnormal ($y_i=1$). The anomaly mask $G_i$ provides pixel-wise annotations for $x_i$, where each pixel has a value of $0$ if normal and $1$ if anomalous. We define a collection of test sets as $\mathcal{T}=\{\mathcal{D}_{test}^1,\mathcal{D}_{test}^2,\cdots,\mathcal{D}_{test}^M\}$, where $\mathcal{D}_{test}^j = \{X_{test}^j, Y_{test}^j\}$ corresponds to a test set from one of $M$ different application domains. Each domain contains its own types of anomalies that differ from those seen in the auxiliary training dataset $\mathcal{D}_{train}$.

In this work, we train the GADS model InCTRLv2 using only the auxiliary dataset $ \mathcal{D}_{train} $, such that it generalizes to all test domains in $\mathcal{T}$. During inference, the few-shot sample prompts are a small set of normal images randomly drawn from the target domain, denoted as $\mathcal{P}_{test}=\{p_1, p_2, \cdots,p_K\}$, where $K$ is typically a small number (\eg, $K \ll N$). $\mathcal{P}_{test}$ is not available during training and is not used to update the generalist model parameters in any way. The resulting model is expected to produce both an image-level anomaly score $s(x) \in \mathbb{R}$ and a pixel-level anomaly map $\mathbf{\mathcal{M}}_x \in \mathbb{R}^{h \times w}$. The image-level anomaly score $s(x) \in [0, 1]$ provides a global prediction of whether the image contains any abnormal content, while the pixel-level map $\mathbf{\mathcal{M}}_x \in [0, 1]$ indicates the likelihood of each spatial location being anomalous. Higher values of $s(x)$ or entries in $\mathbf{\mathcal{M}}_x$ correspond to a higher probability of abnormality.

\subsection{VLM Backbone.}
Following InCTRL~\cite{zhu2024generalistanomalydetectionincontext} and previous works~\cite{zhou2023anomalyclip, jeong2023winclip, chen2023april, li2024promptad}, we select CLIP~\cite{li2023clip} as our VLM backbone to enable GADS. CLIP consists of a text encoder $f_t(\cdot)$ and a visual encoder $f_v(\cdot)$, with the image and text representations from these encoders well aligned by pre-training on web-scale text-image data. Typically, the CLIP visual encoder $f_v(\cdot)$ comprises a series of ViT block layers. From the bottom to the top of layers, the CLIP visual encoder gradually learns the visual patterns at different levels of abstraction~\cite{radford2021learning}. 

\section{Methodology}
\subsection{Overview of InCTRLv2}
In this work, we propose InCTRLv2, a novel framework for few-shot Generalist Anomaly Detection and Segmentation (GADS) that augments our earlier in-context residual learning framework (InCTRL) with joint learning of semantic-guided abnormality and normality learning.

As presented in Fig.~\ref{fig:overall}, InCTRLv2 builds upon the in-context residual mechanism introduced in InCTRL, where a query sample is compared with few-shot normal references to model their discrepancies (Sec. \ref{subsec:residual_learning}). This residual-based learning effectively captures how normal and abnormal samples differ in the visual space. However, since this image sample-based reference is performed purely at the visual level and optimized using labeled abnormal samples, it can be restricted to the appearance variations (\eg, texture, illumination, or material) and the abnormal patterns illustrated by these labeled data, which can lead to misinterpretation of domain-specific visual changes as anomalies and unseen abnormal patterns as normal, respectively.
To overcome this limitation, InCTRLv2 extends the single-branch residual learning framework in InCTRL with a dual-branch framework—including one main branch and one auxiliary branch—to enhance the original framework with a semantic-guided joint abnormality and normality learning strategy (Sec. \ref{sec: joint learning}).
This strategy jointly models two complementary perspectives of anomaly perception: i) Discriminative Anomaly Score Learning (DASL) with both normal and abnormal data in the main branch, which learns a semantic-guided abnormality and normality space that supports the classification of query samples from both the abnormality and normality perspectives; and ii) One-class Anomaly Score Learning (OASL) using only the normal data, which learns generalized normality patterns in a semantic space via an auxiliary branch, focusing on detecting anomalies through the lens of normality solely.
Together, these two branches offer a dual semantic perspective: one emphasizing normal-abnormal discriminative power,
while the other emphasizing semantic of being deviated from the normality. 

To be specific, during training, a set of in-context examples is given, with each consisting of a query image $x$ paired with a hand-crafted text prompt, and a set of few-shot normal sample prompts $\mathcal{P}_{I}$, where both $x$ and $\mathcal{P}_{I}$ are randomly sampled from the auxiliary dataset $\mathcal{D}_{train}$. The main branch not only models in-context residuals by comparing a query sample against the few-shot normal set $\mathcal{P}_{I}$, but also incorporates text-prompt–based semantic priors to establish a discriminative decision space between normal and abnormal samples.
In contrast, the auxiliary branch is trained exclusively on normal samples from the auxiliary dataset, enabling the model to distill a compact one-class decision space that captures the intrinsic characteristics of normality.
During inference, InCTRLv2 integrates three complementary signals derived from the main and auxiliary branches, resulting in accurate and robust anomaly detection and segmentation under few-shot settings. 
The corresponding text prompts of $x$ are then encoded as text features using CLIP's text encoder $f_t(\cdot)$. 
Concurrently, for the extended dual-branch framework, some anomaly detection–specific prompt templates, denoted as $\mathcal{P}^n_T$ and $\mathcal{P}^a_T$ respectively, are also encoded by $f_t(\cdot)$ to provide semantic information for both DASL and OASL modules, with the former taking $x$ as the query image while the latter taking $\hat{x}$, sampled only from $\mathcal{D}_{normal} \subset \mathcal{D}_{train}$, as input. 

Below we present these modules in detail.

\subsection{Residual Learning from In-context Normal Image Prompts}\label{subsec:residual_learning}
InCTRLv2 builds upon the in-context residual scoring mechanism introduced in InCTRL, which aims to compare a query sample against a set of few-shot normal references and model their residual in a domain-agnostic manner. Specifically, the in-context residual scoring module models discrepancies at both the image and patch levels: the former captures global in-context residual features, while the latter provides fine-grained local cues. With the supervision across diverse classes, this module effectively transfers residual knowledge across different domains, laying a strong foundation for generalist anomaly detection and segmentation.

\subsubsection{Image-level Residual Learning}
To capture global in-context residual, we introduce an image-level residual learning component to model the higher-level differences between the query image and the few-shot normal prompts. Specifically, we utilize the class token embedding from the final block of the CLIP visual encoder in this component, as it encapsulates high-level semantic information and is commonly employed in image classification tasks. However, CLIP’s pre-training objective focuses on class-level discrimination, which is not aligned with anomaly detection, where normal and abnormal instances often belong to the same class but differ only in subtle, fine-grained ways. To address this, we introduce an adapter layer $\psi(\cdot; \Theta_{\psi})$, parameterized by $\Theta_{\psi}$, to transform class token embeddings into more AD-sensitive representations. As shown in Fig~\ref{fig:in-context-residual-learning} \textbf{Top}, we calculate the image-level in-context residuals by comparing the adapted feature of the query image with a prototypical representation derived from the few-shot normal images. 

Formally, let $f_v(x)\in \mathbb{R}^{d^\prime}$ denote the class token embedding of an input image $x$ from the visual encoder. We first compute a prototypical feature representation from the set of few-shot normal image prompts $\mathcal{P}_{I}$ by averaging their adapted features:
\begin{equation}
    \mathbf{I}_{p} = \frac{1}{K}\sum_{x_{k}^\prime \in \mathcal{P}_{I}}\psi(f_v(x_{k}^\prime);\Theta_\psi),
    \label{eq:patch-level}
\end{equation}
where $\mathbf{I}_{p} \in \mathbb{R}^{d^\prime}$. Then let $\mathbf{I}_{x} = \psi(f_{v}(x);\Theta_\psi)$ be the adapted features of the query image $x$, and the image-level in-context residual features $\mathbf{F}_{x}$ for $x$ are obtained by performing subtraction between two feature embeddings:
\begin{equation}
    \mathbf{F}_{x} = \mathbf{I}_{x} \ominus \mathbf{I}_{p},
    \label{eq:image-level}
\end{equation}
where $\ominus$ denotes element-wise subtraction. The resulting residual feature is subsequently fed to an image-level residual scoring learner $\eta: \mathbf{F}_x \rightarrow \mathbb{R}$, parameterized by $\Theta_\eta$, to produce the image-level in-context residual score $s_I = \eta(\mathbf{F}_{x};\Theta_\eta)$.

\subsubsection{Multi-layer Patch-level Residual Learning}
In addition to global image-level residuals, we also capture fine-grained in-context residuals by comparing the query image with few-shot normal sample prompts at patch level. Specifically, we utilize the patch token embeddings obtained from the multiple selected block layers of visual encoder $f_v(\cdot)$ to model hierarchical patch-level in-context residuals. Formally, assuming $f_v(\cdot)$ consists of $L$ blocks, for a given query image $x$ and a set of few-shot normal sample prompts $\mathcal{P}_{I}$, we extract a sequence of patch token embeddings $\{F_v^l\}_{l=1}^L$ and $\{F_{v^\prime}^l\}_{l=1}^L$, where each $F_{(\cdot)}^l \in \mathbb{R}^{h \times w \times d} $ and $x^\prime \in \mathcal{P}_{I}$ denotes the spatial grid of patch tokens from layer $l$, and $x^\prime \in \mathcal{P}_{I}$. Here, $h$, $w$, and $d$ represent the height, width, and embedding dimension of the feature maps, respectively.

\begin{figure}[ht]
    \centering
    \includegraphics[width=0.42\textwidth]{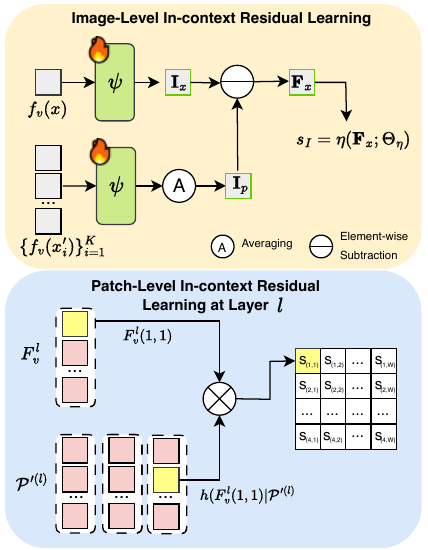}
    \caption{Detailed illustration of the image-level in-context learning and the multi-layer patch-level in-context learning mechanisms in the DASL module. }
    \label{fig:in-context-residual-learning}
\end{figure}

As illustrated in Fig~\ref{fig:in-context-residual-learning} \textbf{Bottom}, at each layer $l$, the patch-level in-context residuals are captured by distances between the embeddings of the query token and the tokens across all few-shot images in $\mathcal{P}_{I}$. Formally, the patch-level residual map is defined as $\mathbf{M}_x^l \in \mathbb{R}^{h \times w}$ where the residual value of each patch at spatial location $(i, j)$ of $x$ is calculated based on its patch embedding and the nearest patch embedding of all images in $\mathcal{P}_{I}$:
\begin{equation}
    \mathbf{M}_x^l(i, j) = 1 - \langle F_v^l(i, j), h(F_{v'}^l(i, j)|\mathcal{P}_{I})\rangle,
    \label{eq:mult-layer-residual-map}
\end{equation}
where $h(F_{v'}^l(i, j)|\mathcal{P}_{I})$ returns the patch token embedding from $\mathcal{P}_{I}$ that is most similar to the query patch $F_v^l(i, j)$ across all image patches, and $\langle \cdot\rangle$ is the cosine similarity function. To obtain a consolidated residual map, we average the residuals across a set of selected transformer layers $\mathcal{L}$. The final patch-level residual map $\mathbf{M}_x \in \mathbb{R}^{h \times w}$ is computed as:
\begin{equation}
    \mathbf{M}_{x} = \frac{1}{|\mathcal{L}|} \sum_{l=1}^\mathcal{L} \mathbf{M}_x^l.
    \label{eq:residual-map}
\end{equation}

Each residual value in $\mathbf{M}_{x}$ can be interpreted as a nearest-neighbor-distance anomaly score, measuring the similarity between a query patch and the closest patch in the prompt set $\mathcal{P}_{I}$. This formulation follows the intuition that anomalous regions are less likely to have close counterparts in normal reference images. Prior studies in patch-based and distance-based anomaly detection~\cite{cohen2020sub,defard2021padim, roth2022towards, pang2015lesinn,pang2018learning} have shown that such nearest-neighbor distances are effective at distinguishing anomalies from normal patterns. Therefore, the resulting residual map $\mathbf{M}_{x}$ serves as a compact, multi-resolution feature representation with strong anomaly-discriminative capacity.

\begin{table}[!ht]
\centering
\resizebox{0.45\textwidth}{!}{
\begin{tabular}{|c|}
\toprule
\textbf{Normal Text Prompt Examples}                                                                      \\ \midrule
\begin{tabular}[c]{@{}c@{}}\textit{‘a photo of a flawless {[}c{]} for visual inspection.’}\\ \textit{‘a cropped photo of a perfect {[}c{]}.’}\\ \textit{‘a blurry photo of the {[}c{]} without defect.’}\\ \textit{‘a dark photo of the unblemished {[}c{]}.’}\\ \textit{‘a jpeg corrupted photo of a {[}c{]} without flaw.’}\end{tabular} \\ \midrule
\textbf{Abnormal Text Prompt Examples}  \\ \midrule
\begin{tabular}[c]{@{}c@{}}\textit{‘a photo of a {[}c{]} with flaw for visual inspection.’}\\ \textit{‘a cropped photo of a {[}c{]} with damage.’}\\ \textit{‘a blurry photo of the {[}c{]} with defect.’}\\ \textit{‘a dark photo of the {[}c{]} with flaw.’}\\ \textit{‘a jpeg corrupted photo of a {[}c{]} with defect.’}\end{tabular}  \\ \bottomrule
\end{tabular}}
\caption{Examples of normal and abnormal text prompts used in InCTRLv2. \textit{[c]} represents a category-level label, \eg, cable.}
\label{example}
\end{table}

\subsection{Joint Learning of Semantic-guided Generalized Abnormality and Normality}
\label{sec: joint learning}
While the in-context residuals capture deviations between a query and normal references, this modeling operates purely within the visual feature space without semantic-level guidance. Consequently, it tends to be influenced by visual variations in the images (\eg, texture, illumination, and material differences). These low-level attributes often vary significantly across domains, leading the model to overfit to domain-specific appearances and misinterpret superficial differences as anomalies. Furthermore, the learning of these residuals relies heavily on both labeled normal and abnormal data from the auxiliary datasets, which can limit the model’s detection ability to the abnormal patterns seen in the auxiliary data.

To address this limitation, we introduce a joint abnormality and normality learning strategy, named Semantic-guided Generalized Abnormality and Normality, which enhances the residual learning module by leveraging the text prompt-based semantic priors encoded in CLIP to jointly model two complementary perspectives of anomaly understanding—one is from discriminative abnormality learning trained with both normal and abnormal data, and the other is from one-class normality learning trained solely on normal data. This strategy is expected to enable the model to move beyond the visual residual space and the normal-abnormal discriminative cues, thereby achieving better generalization capabilities across diverse domains.

To achieve this, we firstly extract text-based semantic features using CLIP’s text encoder $f_t(\cdot)$ based on text prompts of the normal and abnormal classes, which are a set of hand-crafted anomaly detection–specific templates providing semantic context for distinguishing normal and abnormal patterns.
In our work, we adopt the text prompt templates and ensemble strategy proposed in WinCLIP~\cite{jeong2023winclip} as our text prompts, which have demonstrated strong performance in zero- and few-shot anomaly detection (see examples in Table~\ref{example}). 
Formally, let $\mathcal{P}^n_T$ and $\mathcal{P}^a_T$ denote the sets of text prompts for the normal and abnormal categories, respectively. We compute their prototype embeddings as follows:
\begin{equation}
    \mathbf{F}_n = \frac{1}{|\mathcal{P}^n_T|} \sum_{p_i \in \mathcal{P}^n_T} f_t(p_i), \quad 
    \mathbf{F}_a = \frac{1}{|\mathcal{P}^a_T|} \sum_{p_j \in \mathcal{P}^a_T} f_t(p_j).
    \label{eq:text-prompts}
\end{equation}

After that, these semantic prototype embeddings are utilized by the following two semantic-guided generalized abnormality and normality learning modules to model the abnormality from two different perspectives.

\subsubsection{Discriminative Anomaly Score Learning with Both Abnormal and Normal Data}
In contrast to the visual residual space in the original residual learning, the semantic-guided discriminative anomaly score learning (DASL) aims to incorporate the text prompt-based semantic priors to establish a new decision space between normal and abnormal samples, providing complementary semantic cues for anomaly understanding learned from the visual residuals. This is achieved by aligning the visual representations with text prompts that describe normality and abnormality. Specifically, given the normal and abnormal text prototype embeddings extracted above, the visual features of a query image are aligned with these prototypes to compute semantic-guided anomaly scores and maps, which serve as complementary signals to the residual learning-based predictions.

Formally, the semantic-guided anomaly score of the query image $x$ is computed by measuring the cosine similarity between the class token embedding $f_v(x)$ and the text prototype embeddings $\{\mathbf{F}_n, \mathbf{F}_a\}$, which is expressed as:
\begin{equation}
    s_q = \frac{\exp(\langle f_{v}(x), \mathbf{F}_a\rangle) }{\sum_{c \in (a, n)}\exp(\langle f_{v}(x), \mathbf{F}_c \rangle )},
    \label{eq:semantic-AD-score}
\end{equation}
where $c \in (n, a)$ denotes the normal ($n$) and abnormal ($a$) categories.

At the same time, we leverage patch token embeddings of the query image $x$ from the selected ViT block layers to compute a multi-layer semantic-guided anomaly map by aligning them with $\{\mathbf{F}_n, \mathbf{F}_a\}$. Here, we introduce a lightweight adapter $\phi_1(\cdot; \Theta_{\phi_1})$ to i) project patch features to have the same dimensionality of textual embeddings, and ii) provide learnable capacity to adapt the CLIP features toward AD-specific cues. To be specific, at a selected layer $l$, we compute a semantic-guided abnormality-oriented anomaly map $\mathbf{S}_a^l$ for query image $x$ by aligning its projected patch token embeddings $\{\phi_1(F_v^l; \Theta_{\phi_1})\}_{l=1}^L$ with the text prototype embeddings for normality $\mathcal{P}^n_T$ and abnormality $\mathcal{P}^a_T$:
\begin{equation}
    \mathbf{S}_a^l(i, j) = \frac{\exp(\langle\phi_1(\mathbf{F}_v^l(i, j); \Theta_{\phi_1}), \mathbf{F}_a \rangle) }{\sum_{c \in (n, a)}\exp( \langle\phi_1(\mathbf{F}_v^l(i, j); \Theta_{\phi
    _1}), \mathbf{F}_c \rangle)},
    \label{eq:semantic-AD-maps}
\end{equation}
where $\mathbf{F}_v^l(i, j)$ denotes the patch token embedding at location $(i,j)$ from layer $l$. The resulting $\mathbf{S}_a^l(i, j)$ reflects the likelihood of the corresponding patch being anomalous. Similarly, the $l$-th layer semantic-guided normality-oriented anomaly map $\mathbf{S}_n^l$ can be computed by replacing $\mathbf{F}_a$ with $\mathbf{F}_n$ in Eq.~\ref{eq:semantic-AD-maps}. 

To obtain the final semantic-guided anomaly maps, we average the respective layer-wise maps across selected layers $\mathcal{L}$, capturing semantic misalignment signals at different levels of abstraction, which are formulated as:
\begin{equation}
    \mathbf{S}_{a} = \frac{1}{|\mathcal{L}|} \sum_{l \in \mathcal{L}} \mathbf{S}_a^l, \quad
    \mathbf{S}_{n} = \frac{1}{|\mathcal{L}|} \sum_{l \in \mathcal{L}} \mathbf{S}_n^l.
    \label{eq:final-semantic-AD-map}
\end{equation}

In summary, the component produces a semantic-guided anomaly score $s_q$ and semantic-guided anomaly maps $\{\mathbf{S}_n, \mathbf{S}_a\}$, which respectively provide image-level and pixel-level semantic alignment signals with normal and abnormal text prototypes. 

The final image-level anomaly score is obtained by combining the in-context residual score $s_I$ with the semantic-guided anomaly score $s_q$, which can be defined as:
\begin{equation}
    s(x) = (1-\alpha) \frac{s_I + s_q}{2} + \alpha s_p(x),
    \label{eq:finalscore}
\end{equation}
where $s_p(x)=\max(\mathbf{M}_{x})$ is a maximum residual score from the in-context residual anomaly map. $s_p(x)$ is added into Eq. \ref{eq:finalscore} because such patch-level residual scores are crucial for detecting local abnormal regions which the holistic anomaly score $\frac{s_I + s_q}{2}$ can often overlook. $\alpha$ is a hyper-parameter that modulates the contribution of the patch-level residual score. The image-level anomaly score $s(x)$ is then optimized by minimizing the following loss:
\begin{equation}
    \mathcal{L}_I = \frac{1}{N}\sum_{x\in X_{train}}\mathcal{L}_{b}(s(x), y_x),
    \label{eq:global}
\end{equation}
where $\mathcal{L}_{b}$ is specified by a focal loss function due to the class imbalance in $X_{train}$. 

For pixel-level supervision, we consider both semantic-guided anomaly maps $\{\mathbf{S}_n, \mathbf{S}_a\}$ and the in-context residual map $\mathbf{M}_{x}$. The pixel-level anomaly map $\mathbf{M}_{p}$ is obtained by averaging $\mathbf{M}_{x}$ and $\mathbf{S}_a$, which can be fomulated as:
\begin{equation}
    \mathbf{M}_{p} = \frac{1}{2} (\mathbf{M}_{x} \oplus \mathbf{S}_a),
    \label{eq:3}
\end{equation}
where $\oplus$ indicates the element-wise addition. The learning objective for optimizing pixel-level anomaly detection in this component is then defined as:
\begin{equation}
\begin{split}
    \mathcal{L}_{P} & = \frac{1}{N}\sum_{x\in X_{train}}\mathcal{L}_{Focal}([\Phi(\mathbf{S}_n), \Phi(\mathbf{S}_a)], G_x)\\
    & + \mathcal{L}_{Dice}(\Phi(\mathbf{S}_a), G_x) + \mathcal{L}_{Dice}(\Phi(\mathbf{M}_p), G_x),
    \label{eq:local}
\end{split}
\end{equation}
where the $\Phi(\cdot)$ is a reshape and interpolation function that transforms the patch-level anomaly maps into a pixel-level segmentation map, and $G_x$ is the corresponding pixel-level ground-truth map. The $\mathcal{L}_{Focal}(\cdot)$ and $\mathcal{L}_{Dice}(\cdot)$ denote the focal loss~\cite{lin2018focallossdenseobject} and dice loss~\cite{li2020dicelossdataimbalancednlp}, respectively.

Finally, the semantic-guided DASL module is trained using a joint learning objective that integrates both image- and pixel-level losses:
\begin{equation}
    \mathcal{L}_{DASL} = \mathcal{L}_{P} + \mathcal{L}_{I}.
    \label{eq:loss-DASL}
\end{equation}

\subsubsection{One-class Anomaly Score Learning with Solely Normal Data} 
Inspired by previous one-class AD approaches~\cite{ruff2018deep}, we observe that in cross-domain settings, the definition of abnormality often varies drastically across domains (e.g., stain in wood vs. crack in capsule), making decision boundaries learned from both normal and abnormal samples sensitive to domain shifts. In contrast, normal patterns tend to exhibit consistent statistical properties across domains, such as smoothness, texture regularity, and physical plausibility. To leverage this stability, we introduce a semantic-guided one-class anomaly score learning (OASL) to model this transferable normality. This component is trained exclusively on normal samples in the auxiliary data, enabling the model to distill a compact one-class decision space that captures the intrinsic characteristics of normality—a stable and generalizable source of AD patterns across diverse domains.

Formally, the semantic-guided OASL module is trained solely on normal samples from the auxiliary dataset, denoted as $\mathcal{D}_{normal} = \{X_{normal}, Y_{normal}\} \subset \mathcal{D}_{train}$. For a normal sample $\hat{x} \in \mathcal{D}_{normal}$, let $\hat{\mathbf{F}}_v^l$ denote its patch token embeddings extracted from layer $l$ of the visual encoder. This component follows a structure similar to the DASL module but employs another independent lightweight adapter $\phi_2(\cdot; \Theta{\phi_2})$. 

Based on this setup, this component generates the $l$-th layer semantic anomaly map by measuring the cosine similarity between the adapted patch embeddings $\phi_2(\hat{\mathbf{F}}v^l; \Theta{\phi_2})$ and the text prototype embeddings $\{\mathbf{F}_n, \mathbf{F}_a\}$:
\begin{equation}
    \hat{\mathbf{S}}_a^l(i, j) = \frac{\exp(\langle \phi_2(\hat{\mathbf{F}}_v^l(i, j); \Theta{\phi_2}), \mathbf{F}_a\rangle)}{\sum_{c \in (n, a)}\exp(\langle \phi_2(\hat{\mathbf{F}}_v^l(i, j); \Theta{\phi_2}), \mathbf{F}_c \rangle)},
    \label{eq:normality-anomaly-map}
\end{equation}
where $(i, j)$ indicates the spatial location of a patch in $\hat{\mathbf{F}}_v^l$. Similarly, we obtain the $l$-th layer semantic map for the normal category, denoted as $\hat{\mathbf{S}}_n^l$, by replacing $\mathbf{F}_a$ with $\mathbf{F}_n$ in Eq.~\ref{eq:normality-anomaly-map}. The final semantic-guided one-class anomaly maps, $\hat{\mathbf{S}}_{n}$ and $\hat{\mathbf{S}}_{a}$, are then computed by averaging their respective layer-wise maps over the selected layers $\mathcal{L}$, following the same formulation as Eq.~\ref{eq:final-semantic-AD-map}:
\begin{equation}
    \hat{\mathbf{S}}_{a} = \frac{1}{|\mathcal{L}|} \sum_{l \in \mathcal{L}} \hat{\mathbf{S}}_a^l, \quad
    \hat{\mathbf{S}}_{n} = \frac{1}{|\mathcal{L}|} \sum_{l \in \mathcal{L}} \hat{\mathbf{S}}_n^l.
    \label{eq:final-semantic-AD-map-OASL}
\end{equation}

We further integrate the abnormality-oriented map $\hat{\mathbf{S}}_{a}$ with the in-context residual anomaly map $\mathbf{M}_{x}$ to derive the final pixel-level anomaly map $\mathbf{M}_{n}$ in this component, following the same formulation as Eq.~\ref{eq:3} as 
$\mathbf{M}_{n} = \frac{1}{2} (\mathbf{M}_{x} \oplus \hat{\mathbf{S}}_a)$. The training objective of the semantic-guided OASL module is then defined as:
\begin{equation}
\begin{split}
    \mathcal{L}_{OASL} & = \frac{1}{N}\sum_{x\in X_{normal}}\mathcal{L}_{Focal}([\Phi(\hat{\mathbf{S}}_n), \Phi(\hat{\mathbf{S}}_a)], G_x)\\
    & +
    \mathcal{L}_{Dice}(\Phi(\hat{\mathbf{S}}_a), G_x) + \mathcal{L}_{Dice}(\Phi(\mathbf{M}_n), G_x).
    \label{eq:loss-OASL}
\end{split}
\end{equation}

Importantly, the semantic-guided OASL operates independently and in parallel with the DASL module. Their parameters, $\Theta_{\phi_1}$ and $\Theta_{\phi_2}$, are optimized separately to ensure that each component learns different aspects of semantic-guided knowledge without mutual interference with the DASL module. This design enables the two learning processes to provide complementary semantic-guided signals, thereby enriching the overall anomaly understanding.

\subsection{Inference}

During inference, for a given test image $x'$ and the $K$-shot normal sample prompts $\mathcal{P'}$ from the target dataset, they are fed forward through the semantic-guided discriminative and one-class anomaly score learning modules of InCTRLv2, respectively obtaining anomaly maps $\mathbf{M}'_{p}$ and $\mathbf{M}'_{n}$. The final predicted pixel-level anomaly map for anomaly localization is calculated as:
\begin{equation}
    \mathbf{M'} = (1 - \beta) \Phi(\mathbf{M}'_{p}) +\beta \Phi(\mathbf{M}'_{n}),
    \label{eq:final-map}
\end{equation}
where $\beta$ is a hyper-parameter. Lastly, we obtain the final anomaly score of $s(x')$ via Eq. \ref{eq:finalscore}

\section{Experiments}
\subsection{Experimental Setup}
\subsubsection{Datasets.}
To assess the zero-shot capability of InCTRLv2, We conduct evaluation on ten real-world anomaly detection datasets, including industrial defective AD datasets (VisA~\cite{zou2022spot}, MVTecAD~\cite{bergmann2019mvtec}, ELPV~\cite{deitsch2019automatic}, SDD~\cite{tabernik2020segmentation}, AITEX~\cite{silvestre2019public}), medical AD datasets (BrainMRI~\cite{salehi2021multiresolution}, HeadCT~\cite{salehi2021multiresolution}, BraTS~\cite{menze2014multimodal}), and semantic AD datasets (MNIST~\cite{lecun1998gradient}, CIFAR10~\cite{krizhevsky2009cifar}). For semantic anomalies, we consider both one-vs-all and multi-class protocols~\cite{Cao_2023_ICCV,ruff2020deep, zhu2024generalistanomalydetectionincontext}. Under the one-vs-all protocol, one class is used as normal, with the other classes treated as abnormal; while under the multi-class protocol, images of even-number classes from MNIST and animal-related classes from CIFAR-10 are treated as normal, with the images of the other classes considered as anomalies. 

We follow a cross-dataset evaluation protocol. Specifically, the test split of MVTec AD is used as the auxiliary training data for the main branch, while the training split of MVTec AD serves as the auxiliary data for the OASL branch, on which the AD models are trained. The trained models are then evaluated on the test sets of the other eight datasets without any further training, thereby assessing their generalization ability across domains. For evaluating performance on MVTec AD itself, we follow the same protocol as VisA, where training is conducted on the VisA and testing is performed on the MVTec AD test set. 

The few-shot normal prompts for the target data are randomly sampled from the training set of target datasets and remain the same for all models for fair comparison. We evaluate the performance with the number of few-shot normal prompt set to $K = 1, 2, 4$. 
The reported results are averaged over three independent runs with different random seeds. 

\subsubsection{Competing Methods and Evaluation Metrics.}
InCTRLv2 is compared with three conventional full-shot AD approaches, including SPADE~\cite{cohen2020sub}, PaDiM~\cite{defard2021padim}, and PatchCore~\cite{roth2022towards}, all of which are adapted to the few-shot setting by performing their distance-based anomaly scoring based on the few-shot normal samples. We also compare against state-of-the-art (SotA) VLM-based methods, including WinCLIP~\cite{jeong2023winclip} and InCTRL~\cite{zhu2024generalistanomalydetectionincontext} (with its ViT-L/14 variant, denoted as InCTRL+), both of which adopt CLIP as their backbone, and ResAD~\cite{yao2024resad}, which is built upon ImageBind~\cite{girdhar2023imagebind}. 

As for evaluation metrics, following previous works~\cite{jeong2023winclip, yao2024resad, zhu2024generalistanomalydetectionincontext}, we use two popular metrics: Area Under the Receiver Operating Characteristic (AUROC) and Average Precision (AP) to assess the image-level AD performance; for pixel-level AD performance, we employ AUROC and Area Under Per Region Overlap (PRO) to provide a more detailed analysis. 

\subsubsection{Implementation Details.}
By default, all CLIP-based models, including WinCLIP~\cite{jeong2023winclip}, InCTRL~\cite{zhu2024generalistanomalydetectionincontext} and InCTRLv2, the same CLIP implementation, OpenCLIP~\cite{ilharco2021openclip}, and its public pre-trained backbones ViT-B/16+ or ViT-L/14 are used in experiments. Our data preprocessing aligns with OpenCLIP across all datasets. Specifically, this involves channel-wise standardization using a predefined mean and standard deviation after scaling RGB images to the range of [0, 1], followed by bicubic resizing based on Pillow library. In addition, we resize the input resolution to 240$\times$240 to match ViT-B/16+ and 518$\times$518 to match ViT-L/14. Adam is used as the optimizer and the initial learning rate is set to 1e-3 by default. The text prompts used in InCTRLv2 are kept exactly the same as WinCLIP, InCTRL and ResAD for fair comparison. To enable the model to recognize both normal and abnormal objects while preventing overfitting, the training epochs are set to 10 with a batch size of 48 on a single GPU (NVIDIA GeForce RTX 3090). 
We implement SPADE, PaDiM and WinCLIP to use the same image prompts as InCTRLv2 for fair comparison, and adopt the official implementation of PatchCore, InCTRL and ResAD. 

\subsubsection{Data Availability.}
The related code and data for InCTRLv2 are available at \renewcommand\UrlFont{\color{blue}}\url{https://github.com/mala-lab/InCTRLv2}.

\begin{table*}[ht]
\setlength{\tabcolsep}{4pt}
\centering
\resizebox{\textwidth}{!}{
\begin{tabular}{c|c|c|ccc|ccccc}
\toprule
\multirow{2}{*}{\textbf{Setup}} & \multirow{2}{*}{\textbf{Domain}}  & \multirow{2}{*}{\textbf{Dataset}} & \multicolumn{3}{c|}{\textbf{Non-VLM-based Methods}}   & \multicolumn{5}{c}{\textbf{VLM-based Methods}}                                                                     \\
                                  &                                      &                                   & \textbf{SPADE}        & \textbf{PaDiM}       & \textbf{Patchcore}   & \textbf{WinCLIP}      & \textbf{ResAD} & \textbf{InCTRL}      & \textbf{InCTRL+}     & \textbf{InCTRLv2}  \\ \midrule
\multirow{12}{*}{\textbf{1-shot}} & \multirow{5}{*}{\textbf{Industrial}} & \textbf{VisA}                     & (67.2{\tiny±3.8}, 72.1{\tiny±2.9})  & (61.8{\tiny±0.3}, 67.2{\tiny±1.1}) & (75.8{\tiny±2.8}, 79.9{\tiny±1.9}) & (82.5{\tiny±0.1}, 83.5{\tiny±0.1})  & (79.4{\tiny±3.4}, 81.9{\tiny±1.1})   & (84.8{\tiny±0.1}, 85.2{\tiny±1.6}) & ({\color[HTML]{0000FF} \textbf{85.1{\tiny±0.9}}}, {\color[HTML]{0000FF} \textbf{86.6{\tiny±0.9}}}) & ({\color[HTML]{FF0000} \textbf{89.0{\tiny±0.5}}}, {\color[HTML]{FF0000} \textbf{90.9{\tiny±0.5}}}) \\
                                  &                                      & \textbf{MVTecAD}                  & (74.5{\tiny±1.3}, 87.9{\tiny±0.9})  & (76.0{\tiny±0.5}, 87.2{\tiny±0.6}) & (82.8{\tiny±0.8}, 91.7{\tiny±1.0}) & ({\color[HTML]{0000FF} \textbf{92.7{\tiny±0.7}}}, {\color[HTML]{0000FF} \textbf{95.9{\tiny±0.3}}})  & (89.3{\tiny±1.4}, 92.7{\tiny±1.0})   & (92.2{\tiny±1.1}, 95.2{\tiny±0.9}) & (90.6{\tiny±0.5}, 95.2{\tiny±0.6}) & ({\color[HTML]{FF0000} \textbf{94.5{\tiny±0.3}}}, {\color[HTML]{FF0000} \textbf{97.2{\tiny±0.2}}}) \\
                                  &                                      & \textbf{ELPV}                     & (50.0{\tiny±1.0}, 61.0±{\tiny1.0})  & (54.3{\tiny±4.6}, 65.7{\tiny±1.0}) & (69.5{\tiny±2.6}, 82.2{\tiny±1.9}) & (72.0{\tiny±2.6}, 84.4{\tiny±2.2})  & (80.9{\tiny±2.9}, 77.4{\tiny±5.7})   & ({\color[HTML]{0000FF} \textbf{81.1{\tiny±0.4}}}, {\color[HTML]{0000FF} \textbf{89.4{\tiny±1.3}}}) & (80.9{\tiny±1.8}, 88.4{\tiny±3.9}) & ({\color[HTML]{FF0000} \textbf{82.2{\tiny±2.6}}}, {\color[HTML]{FF0000} \textbf{90.7{\tiny±1.6}}}) \\
                                  &                                      & \textbf{SDD}                      & (72.5{\tiny±1.0}, 36.3{\tiny±8.3})  & (71.9{\tiny±1.9}, 34.5{\tiny±1.2}) & (90.2{\tiny±0.6}, 67.1{\tiny±1.0}) & (94.1{\tiny±0.3}, 86.3{\tiny±0.9})  & (91.8{\tiny±1.6}, 79.1{\tiny±1.1})   & (95.9{\tiny±1.1}, {\color[HTML]{0000FF} \textbf{88.5{\tiny±3.8}}}) & ({\color[HTML]{0000FF} \textbf{96.2{\tiny±0.5}}}, {\color[HTML]{0000FF} \textbf{88.5{\tiny±2.3}}}) & ({\color[HTML]{FF0000} \textbf{98.1{\tiny±0.8}}}, {\color[HTML]{FF0000} \textbf{94.9{\tiny±2.0}}}) \\
                                  &                                      & \textbf{AITEX}                    & (70.7{\tiny±1.4}, 44.7{\tiny±4.4})  & (72.4{\tiny±3.8}, 48.0{\tiny±4.4}) & (72.9{\tiny±2.2}, 37.8{\tiny±0.7}) & (72.2{\tiny±3.6}, 48.6{\tiny±5.3})  & ({\color[HTML]{FF0000} \textbf{87.7{\tiny±1.1}}}, {\color[HTML]{FF0000} \textbf{73.5{\tiny±3.8}}})   & (76.9{\tiny±3.7}, 48.7{\tiny±1.7}) & (77.8{\tiny±0.5}, 50.4{\tiny±1.1}) & ({\color[HTML]{0000FF} \textbf{78.6{\tiny±0.5}}}, {\color[HTML]{0000FF} \textbf{52.3{\tiny±0.5}}}) \\ \cmidrule{2-11}
                                  & \multirow{3}{*}{\textbf{Medical}}    & \textbf{BrainMRI}                 & (66.5{\tiny±3.3}, 93.0{\tiny±0.7})  & (65.4{\tiny±2.8}, 90.3{\tiny±2.0}) & (59.7{\tiny±7.5}, 88.3{\tiny±1.7}) & (92.9{\tiny±1.0}, 98.0{\tiny±0.2})  & (82.8{\tiny±4.5}, 96.6{\tiny±1.1})   & (97.3{\tiny±1.2}, {\color[HTML]{0000FF} \textbf{99.3{\tiny±0.2}}}) & ({\color[HTML]{0000FF} \textbf{97.5{\tiny±0.8}}}, {\color[HTML]{FF0000} \textbf{99.6{\tiny±1.6}}}) & ({\color[HTML]{FF0000} \textbf{98.0{\tiny±0.3}}}, {\color[HTML]{FF0000} \textbf{99.6{\tiny±0.1}}}) \\
                                  &                                      & \textbf{HeadCT}                   & (54.4{\tiny±9.4}, 81.0{\tiny±3.0})  & (55.0{\tiny±5.7}, 85.4{\tiny±2.3}) & (57.5{\tiny±7.7}, 84.4{\tiny±1.6}) & (90.6{\tiny±2.9}, 95.7{\tiny±1.4})  & (76.8{\tiny±1.5}, 90.8{\tiny±1.5})   & ({\color[HTML]{0000FF} \textbf{91.2{\tiny±0.6}}}, {\color[HTML]{0000FF} \textbf{96.7{\tiny±0.4}}}) & (89.5{\tiny±1.3}, 96.4{\tiny±1.7}) & ({\color[HTML]{FF0000} \textbf{91.3{\tiny±1.7}}}, {\color[HTML]{FF0000} \textbf{97.3{\tiny±0.8}}}) \\
                                  &                                      & \textbf{BraTS}                    & (67.3{\tiny±4.6}, 92.4{\tiny±1.1})  & (66.4{\tiny±1.1}, 92.6{\tiny±0.4}) & (84.0{\tiny±0.7}, 97.3{\tiny±0.2}) & (65.1{\tiny±0.6}, 92.4{\tiny±0.4})  & (61.5{\tiny±9.1}, 91.0{\tiny±1.8})   & (79.6{\tiny±0.8}, {\color[HTML]{0000FF} \textbf{96.2{\tiny±0.2}}}) & ({\color[HTML]{0000FF} \textbf{82.3{\tiny±1.5}}}, {\color[HTML]{FF0000} \textbf{97.0{\tiny±0.3}}}) & ({\color[HTML]{FF0000} \textbf{85.3{\tiny±1.2}}}, {\color[HTML]{FF0000} \textbf{97.0{\tiny±0.2}}}) \\ \cmidrule{2-11}
                                  & \multirow{4}{*}{\textbf{Semantical}} & \textbf{MNIST$_1$}         & (75.9{\tiny±0.2}, 96.1{\tiny±1.7})  & -                    & (74.4{\tiny±0.8}, 95.5{\tiny±0.3}) & (78.0{\tiny±1.3}, 94.2{\tiny±0.6})  & (72.3{\tiny±5.1}, 94.8{\tiny±0.7})   & ({\color[HTML]{0000FF} \textbf{87.3{\tiny±1.5}}}, {\color[HTML]{0000FF} \textbf{95.7{\tiny±0.7}}}) & (85.4{\tiny±1.1}, 93.3{\tiny±0.9}) & ({\color[HTML]{FF0000} \textbf{87.7{\tiny±0.8}}}, {\color[HTML]{FF0000} \textbf{96.0{\tiny±0.3}}}) \\
                                  &                                      & \textbf{MNIST$_2$}             & (52.6{\tiny±4.8}, 51.8{\tiny±2.8})  & -                    & (41.3{\tiny±3.6}, 44.7{\tiny±2.0}) & ({\color[HTML]{0000FF} \textbf{63.1{\tiny±0.2}}}, {\color[HTML]{FF0000} \textbf{61.0{\tiny±0.4}}})  & (40.4{\tiny±0.9}, 58.0{\tiny±9.3})   & (60.2{\tiny±1.1}, 58.6{\tiny±1.3}) & (61.3{\tiny±3.3}, 59.9{\tiny±3.1}) & ({\color[HTML]{FF0000} \textbf{63.6{\tiny±2.2}}}, {\color[HTML]{0000FF} \textbf{60.7{\tiny±1.8}}}) \\
                                  &                                      & \textbf{CIFAR10$_1$}        & (81.7{\tiny±1.4}, 96.3{\tiny±0.9})  & -                    & (58.5{\tiny±0.4}, 91.3{\tiny±0.2}) & ({\color[HTML]{0000FF} \textbf{92.5{\tiny±0.2}}}, {\color[HTML]{0000FF} \textbf{99.0{\tiny±0.1}}})  & (70.5{\tiny±5.4}, 94.3{\tiny±0.8})   & (91.3{\tiny±0.2}, 98.6{\tiny±0.4}) & (91.9{\tiny±0.7}, 98.0{\tiny±0.5}) & ({\color[HTML]{FF0000} \textbf{97.2{\tiny±0.4}}}, {\color[HTML]{FF0000} \textbf{99.7{\tiny±0.2}}}) \\
                                  &                                      & \textbf{CIFAR10$_2$}           & (60.7{\tiny±3.0}, 47.8{\tiny±3.5})  & -                    & (70.5{\tiny±1.6}, 57.5{\tiny±1.3}) & ({\color[HTML]{0000FF} \textbf{91.3{\tiny±0.4}}}, {\color[HTML]{0000FF} \textbf{86.5{\tiny±1.8}}})  & (60.9{\tiny±11.8}, 64.7{\tiny±1.5})   & (89.8{\tiny±0.9}, 84.1{\tiny±1.8}) & (90.1{\tiny±3.9}, 79.8{\tiny±2.6}) & ({\color[HTML]{FF0000} \textbf{96.6{\tiny±1.3}}}, {\color[HTML]{FF0000} \textbf{93.1{\tiny±0.5}}}) \\ \midrule
\multirow{12}{*}{\textbf{2-shot}} & \multirow{5}{*}{\textbf{Industrial}} & \textbf{VisA}                     & (79.5{\tiny±4.5}, 81.8{\tiny±3.1})  & (68.0{\tiny±4.2}, 81.8{\tiny±3.1}) & (81.7{\tiny±2.8}, 84.1{\tiny±2.3}) & (84.2{\tiny±2.4}, 85.9{\tiny±2.1})  & (85.6{\tiny±1.6}, 86.2{\tiny±2.6})   & (85.8{\tiny±2.2}, 87.7{\tiny±1.6}) & ({\color[HTML]{0000FF} \textbf{87.1{\tiny±0.7}}}, {\color[HTML]{0000FF} \textbf{87.8{\tiny±0.6}}}) & ({\color[HTML]{FF0000} \textbf{90.9{\tiny±0.2}}}, {\color[HTML]{FF0000} \textbf{92.3{\tiny±0.1}}}) \\
                                  &                                      & \textbf{MVTecAD}                  & (81.7{\tiny±5.4}, 92.2{\tiny±2.3})  & (78.5{\tiny±2.5}, 89.0{\tiny±1.5}) & (85.8{\tiny±3.4}, 93.9{\tiny±1.2}) & (93.1{\tiny±1.9}, 96.5{\tiny±0.7})  & (92.3{\tiny±1.7}, 94.8{\tiny±2.3})   & ({\color[HTML]{0000FF} \textbf{94.0{\tiny±1.5}}}, {\color[HTML]{0000FF} \textbf{96.9{\tiny±0.4}}}) & (92.5{\tiny±0.4}, {\color[HTML]{0000FF} \textbf{96.9{\tiny±0.2}}}) & ({\color[HTML]{FF0000} \textbf{95.7{\tiny±0.5}}}, {\color[HTML]{FF0000} \textbf{97.9{\tiny±0.2}}}) \\
                                  &                                      & \textbf{ELPV}                     & (51.7{\tiny±1.2}, 61.8{\tiny±0.7})  & (59.4{\tiny±8.3}, 70.7{\tiny±5.8}) & (71.6{\tiny±3.1}, 84.0{\tiny±3.1}) & (72.6{\tiny±2.0}, 84.9{\tiny±1.0})  & (82.9{\tiny±0.6}, 82.2{\tiny±2.2})   & ({\color[HTML]{0000FF} \textbf{83.9{\tiny±0.3}}}, {\color[HTML]{0000FF} \textbf{91.3{\tiny±0.8}}}) & (82.6{\tiny±1.2}, 90.2{\tiny±0.8}) & ({\color[HTML]{FF0000} \textbf{84.1{\tiny±0.2}}}, {\color[HTML]{FF0000} \textbf{91.4{\tiny±0.5}}}) \\
                                  &                                      & \textbf{SDD}                      & (72.9{\tiny±4.1}, 36.6{\tiny±10.5}) & (72.1{\tiny±1.5}, 33.7{\tiny±0.8}) & (90.2{\tiny±0.6}, 67.6{\tiny±0.3}) & (94.2{\tiny±0.6}, 86.5{\tiny±0.4})  & (93.8{\tiny±0.3}, 80.4{\tiny±0.3})   & (97.2{\tiny±1.1}, {\color[HTML]{0000FF} \textbf{91.7{\tiny±0.9}}}) & ({\color[HTML]{0000FF} \textbf{97.4{\tiny±1.8}}}, 91.1{\tiny±3.6}) & ({\color[HTML]{FF0000} \textbf{98.5{\tiny±0.3}}}, {\color[HTML]{FF0000} \textbf{95.5{\tiny±0.4}}}) \\
                                  &                                      & \textbf{AITEX}                    & (72.7{\tiny±0.4}, 47.0{\tiny±0.8})  & (78.4{\tiny±2.8}, 52.9{\tiny±3.4}) & (73.9{\tiny±1.7}, 37.8{\tiny±0.8}) & (72.6{\tiny±5.5}, 50.0{\tiny±4.3})  & ({\color[HTML]{FF0000} \textbf{83.1{\tiny±4.4}}}, {\color[HTML]{FF0000} \textbf{67.3{\tiny±3.5}}})   & (76.1{\tiny±2.9}, 51.9{\tiny±2.2}) & (79.6{\tiny±2.0}, 56.0{\tiny±1.8}) & ({\color[HTML]{0000FF} \textbf{80.6{\tiny±4.6}}}, {\color[HTML]{0000FF} \textbf{56.8{\tiny±4.5}}}) \\ \cmidrule{2-11}
                                  & \multirow{3}{*}{\textbf{Medical}}    & \textbf{BrainMRI}                 & (75.4{\tiny±4.8}, 95.2{\tiny±0.9})  & (65.7{\tiny±2.2}, 90.2{\tiny±4.6}) & (70.6{\tiny±0.9}, 92.1{\tiny±1.7}) & (93.4{\tiny±1.2}, 98.9{\tiny±0.3})  & (84.1{\tiny±2.4}, 93.1{\tiny±1.2})   & (97.3{\tiny±2.7}, {\color[HTML]{0000FF} \textbf{99.4{\tiny±1.3}}}) & ({\color[HTML]{0000FF} \textbf{97.7{\tiny±0.5}}}, 98.6{\tiny±0.4}) & ({\color[HTML]{FF0000} \textbf{98.1{\tiny±0.1}}}, {\color[HTML]{FF0000} \textbf{99.7{\tiny±0.0}}}) \\
                                  &                                      & \textbf{HeadCT}                   & (64.5{\tiny±3.4}, 85.1{\tiny±2.2})  & (59.5{\tiny±3.6}, 87.6{\tiny±1.7}) & (73.6{\tiny±9.6}, 91.3{\tiny±0.2}) & (91.5{\tiny±1.5}, 97.5{\tiny±1.2})  & (75.5{\tiny±0.7}, 90.9{\tiny±0.8})   & ({\color[HTML]{0000FF} \textbf{92.9{\tiny±2.5}}}, {\color[HTML]{0000FF} \textbf{98.1{\tiny±1.3}}}) & (90.1{\tiny±2.1}, 96.9{\tiny±1.0}) & ({\color[HTML]{FF0000} \textbf{93.5{\tiny±1.0}}}, {\color[HTML]{FF0000} \textbf{98.3{\tiny±0.3}}}) \\
                                  &                                      & \textbf{BraTS}                    & (67.5{\tiny±8.9}, 93.0{\tiny±2.6})  & (67.3{\tiny±3.8}, 93.0{\tiny±1.4}) & (84.9{\tiny±0.7}, 97.5{\tiny±0.2}) & (67.6{\tiny±1.7}, 92.3{\tiny±0.4})  & (78.4{\tiny±3.9}, 94.3{\tiny±0.7})   & (83.2{\tiny±2.5}, 96.8{\tiny±0.4}) & ({\color[HTML]{0000FF} \textbf{84.0{\tiny±1.3}}}, {\color[HTML]{0000FF} \textbf{97.1{\tiny±0.3}}}) & ({\color[HTML]{FF0000} \textbf{87.7{\tiny±0.8}}}, {\color[HTML]{FF0000} \textbf{97.6{\tiny±0.2}}}) \\  \cmidrule{2-11}
                                  & \multirow{4}{*}{\textbf{Semantical}} & \textbf{MNIST$_1$}         & (77.9{\tiny±2.4}, 96.5{\tiny±0.4})  & -                    & (75.6{\tiny±0.4}, 95.6{\tiny±0.1}) & (81.0{\tiny±0.8}, 96.3{\tiny±0.1})  & (80.0{\tiny±2.5}, 95.8{\tiny±0.5})   & ({\color[HTML]{0000FF} \textbf{89.2{\tiny±0.9}}}, {\color[HTML]{0000FF} \textbf{97.5{\tiny±0.4}}}) & (86.7{\tiny±0.5}, 94.2{\tiny±0.2}) & ({\color[HTML]{FF0000} \textbf{89.5{\tiny±0.3}}}, {\color[HTML]{FF0000} \textbf{97.7{\tiny±0.2}}}) \\
                                  &                                      & \textbf{MNIST$_2$}             & (59.5{\tiny±6.0}, 61.5{\tiny±6.8})  & -                    & (47.3{\tiny±4.1}, 48.2{\tiny±2.5}) & (63.2{\tiny±0.0}, 61.4{\tiny±0.5})  & (54.4{\tiny±4.8}, 60.4{\tiny±7.6})   & ({\color[HTML]{0000FF} \textbf{63.5{\tiny±1.0}}}, {\color[HTML]{0000FF} \textbf{61.8{\tiny±1.2}}}) & (63.0{\tiny±1.9}, 61.6{\tiny±2.7}) & ({\color[HTML]{FF0000} \textbf{66.0{\tiny±1.4}}}, {\color[HTML]{FF0000} \textbf{64.1{\tiny±1.8}}}) \\
                                  &                                      & \textbf{CIFAR10$_1$}        & (82.3{\tiny±1.4}, 97.1{\tiny±0.3})  & -                    & (60.3{\tiny±0.9}, 92.6{\tiny±0.2}) & (92.5{\tiny±0.1}, 99.0{\tiny±0.1})  & (71.6{\tiny±5.5}, 94.4{\tiny±0.7})   & (93.5{\tiny±0.2}, {\color[HTML]{0000FF} \textbf{99.2{\tiny±0.0}}}) & ({\color[HTML]{0000FF} \textbf{94.1{\tiny±0.7}}}, 98.3{\tiny±0.3}) & ({\color[HTML]{FF0000} \textbf{97.6{\tiny±0.4}}}, {\color[HTML]{FF0000} \textbf{99.7{\tiny±0.1}}}) \\
                                  &                                      & \textbf{CIFAR10$_2$}           & (65.5{\tiny±4.2}, 50.2{\tiny±3.5})  & -                    & (70.3{\tiny±0.8}, 57.4{\tiny±1.5}) & (91.4{\tiny±0.5}, 87.6{\tiny±1.6})  & (76.4{\tiny±1.0}, 65.7{\tiny±2.9})   & (92.4{\tiny±0.5}, {\color[HTML]{0000FF} \textbf{89.9{\tiny±1.0}}}) & ({\color[HTML]{0000FF} \textbf{92.5{\tiny±2.5}}}, 84.7{\tiny±2.3}) & ({\color[HTML]{FF0000} \textbf{97.1{\tiny±2.0}}}, {\color[HTML]{FF0000} \textbf{93.9{\tiny±1.6}}}) \\ \midrule
\multirow{12}{*}{\textbf{4-shot}} & \multirow{5}{*}{\textbf{Industrial}} & \textbf{VisA}                     & (81.1{\tiny±4.0}, 82.6{\tiny±2.4})  & (73.5{\tiny±3.1}, 75.8{\tiny±1.8}) & (84.3{\tiny±2.5}, 86.0{\tiny±1.6}) & (87.3{\tiny±1.8}, 88.8{\tiny±1.8}) & (87.3{\tiny±1.7}, 88.2{\tiny±3.0})   & (87.7{\tiny±1.9}, {\color[HTML]{0000FF} \textbf{90.2{\tiny±2.7}}}) & ({\color[HTML]{0000FF} \textbf{89.0{\tiny±0.5}}}, 89.9{\tiny±0.6}) & ({\color[HTML]{FF0000} \textbf{92.0{\tiny±0.3}}}, {\color[HTML]{FF0000} \textbf{93.1{\tiny±0.2}}}) \\
                                  &                                      & \textbf{MVTecAD}                  & (82.8{\tiny±4.4}, 92.4{\tiny±1.5})  & (80.5{\tiny±1.8}, 90.9{\tiny±1.3}) & (88.5{\tiny±2.6}, 95.0{\tiny±1.3}) & (94.0{\tiny±2.1}, 96.8{\tiny±0.8})  & ({\color[HTML]{0000FF} \textbf{95.6{\tiny±2.0}}}, 96.5{\tiny±2.7})   & (94.5{\tiny±1.8}, 97.2{\tiny±0.6}) & (93.2{\tiny±0.5}, {\color[HTML]{0000FF} \textbf{97.3{\tiny±0.5}}}) & ({\color[HTML]{FF0000} \textbf{96.4{\tiny±0.6}}}, {\color[HTML]{FF0000} \textbf{98.1{\tiny±0.2}}}) \\
                                  &                                      & \textbf{ELPV}                     & (53.7{\tiny±1.3}, 62.7{\tiny±1.1})  & (61.2{\tiny±8.0}, 72.4{\tiny±6.7}) & (75.6{\tiny±7.3}, 87.1{\tiny±4.2}) & (75.4{\tiny±0.9}, 86.4{\tiny±0.4})  & ({\color[HTML]{0000FF} \textbf{84.5{\tiny±0.9}}}, 87.1{\tiny±2.8})   & ({\color[HTML]{FF0000} \textbf{84.6{\tiny±1.1}}}, {\color[HTML]{0000FF} \textbf{91.4{\tiny±0.9}}}) & (83.8{\tiny±2.5}, 90.6{\tiny±3.1}) & ({\color[HTML]{FF0000} \textbf{84.6{\tiny±1.6}}}, {\color[HTML]{FF0000} \textbf{91.6{\tiny±2.7}}}) \\
                                  &                                      & \textbf{SDD}                      & (73.1{\tiny±2.0}, 38.5{\tiny±1.8})  & (74.2{\tiny±1.4}, 35.1{\tiny±1.2}) & (92.3{\tiny±0.8}, 70.3{\tiny±1.3}) & (94.3{\tiny±0.4}, 86.8{\tiny±0.3})  & (92.2{\tiny±0.3}, 78.1{\tiny±0.8})   & (97.5{\tiny±0.6}, 92.4{\tiny±1.5}) & ({\color[HTML]{0000FF} \textbf{98.3{\tiny±0.9}}}, {\color[HTML]{0000FF} \textbf{93.8{\tiny±0.9}}}) & ({\color[HTML]{FF0000} \textbf{98.9{\tiny±0.2}}}, {\color[HTML]{FF0000} \textbf{96.7{\tiny±0.5}}}) \\
                                  &                                      & \textbf{AITEX}                    & (71.8{\tiny±1.1}, 45.1{\tiny±3.1})  & (78.7{\tiny±3.8}, 54.0{\tiny±5.3}) & (73.3{\tiny±0.2}, 37.7{\tiny±0.1}) & (76.4{\tiny±2.5}, 51.3{\tiny±1.7})  & ({\color[HTML]{FF0000} \textbf{86.9{\tiny±0.4}}}, {\color[HTML]{FF0000} \textbf{72.0{\tiny±1.1}}})   & (79.0{\tiny±1.8}, 54.8{\tiny±1.6}) & (81.2{\tiny±0.5}, 56.5{\tiny±1.0}) & ({\color[HTML]{0000FF} \textbf{82.6{\tiny±3.9}}}, {\color[HTML]{0000FF} \textbf{63.4{\tiny±7.1}}}) \\  \cmidrule{2-11}
                                  & \multirow{3}{*}{\textbf{Medical}}    & \textbf{BrainMRI}                 & (75.9{\tiny±7.0}, 95.8{\tiny±1.7})  & (79.2{\tiny±4.8}, 95.6{\tiny±1.1}) & (79.4{\tiny±4.0}, 94.5{\tiny±1.7}) & (94.1{\tiny±0.2},  99.0{\tiny±0.1}) & (79.7{\tiny±2.5}, 93.3{\tiny±0.8})   & (97.5{\tiny±1.6}, {\color[HTML]{0000FF} \textbf{99.4{\tiny±1.3}}}) & ({\color[HTML]{0000FF} \textbf{97.8{\tiny±2.7}}}, 98.8{\tiny±0.9}) & ({\color[HTML]{FF0000} \textbf{98.4{\tiny±0.1}}}, {\color[HTML]{FF0000} \textbf{99.7{\tiny±0.1}}}) \\
                                  &                                      & \textbf{HeadCT}                   & (62.4{\tiny±1.2}, 85.4{\tiny±1.6})  & (62.2{\tiny±1.3}, 89.0{\tiny±1.1}) & (80.5{\tiny±0.6}, 94.1{\tiny±0.9}) & (91.2{\tiny±0.3}, {\color[HTML]{0000FF} \textbf{97.4{\tiny±0.2}}})  & (85.5{\tiny±1.1}, 91.7{\tiny±1.9})   & ({\color[HTML]{0000FF} \textbf{93.3{\tiny±1.3}}}, {\color[HTML]{FF0000} \textbf{98.4{\tiny±1.1}}}) & (91.7{\tiny±0.9}, 97.3{\tiny±0.2}) & ({\color[HTML]{FF0000} \textbf{93.7{\tiny±0.4}}}, {\color[HTML]{FF0000} \textbf{98.4{\tiny±0.4}}}) \\
                                  &                                      & \textbf{BraTS}                    & (70.5{\tiny±3.0}, 93.3{\tiny±1.0})  & (68.5{\tiny±1.5}, 92.5{\tiny±0.6}) & (86.9{\tiny±1.5}, 97.6{\tiny±0.3}) & (68.5{\tiny±0.9}, 92.8{\tiny±0.4})  & (81.5{\tiny±2.1}, 95.4{\tiny±1.2})   & ({\color[HTML]{0000FF} \textbf{87.4{\tiny±1.9}}}, {\color[HTML]{0000FF} \textbf{97.5{\tiny±0.6}}}) & (87.0{\tiny±0.4}, 97.3{\tiny±0.2}) & ({\color[HTML]{FF0000} \textbf{89.5{\tiny±0.1}}}, {\color[HTML]{FF0000} \textbf{98.0{\tiny±0.2}}}) \\  \cmidrule{2-11}
                                  & \multirow{4}{*}{\textbf{Semantical}} & \textbf{MNIST$_1$}         & (81.0{\tiny±0.9}, 96.6{\tiny±0.8})  & -                    & (83.3{\tiny±0.9}, 97.2{\tiny±0.2}) & (85.1{\tiny±1.0}, 97.1{\tiny±0.2})  & (81.6{\tiny±6.4}, 96.4{\tiny±0.8})   & ({\color[HTML]{0000FF} \textbf{90.2{\tiny±1.6}}}, {\color[HTML]{0000FF} \textbf{98.0{\tiny±0.7}}}) & (87.6{\tiny±1.2}, 95.8{\tiny±1.5}) & ({\color[HTML]{FF0000} \textbf{90.4{\tiny±0.9}}}, {\color[HTML]{FF0000} \textbf{98.3{\tiny±0.3}}}) \\
                                  &                                      & \textbf{MNIST$_2$}             & (58.8{\tiny±4.1}, 61.1{\tiny±5.3})  & -                    & (49.7{\tiny±4.4}, 50.4{\tiny±2.5}) & (63.2{\tiny±0.4}, 61.1{\tiny±1.1})  & (46.5{\tiny±3.8}, 60.7{\tiny±1.7})   & ({\color[HTML]{0000FF} \textbf{64.3{\tiny±0.7}}}, {\color[HTML]{0000FF} \textbf{62.0{\tiny±0.9}}}) & (63.4{\tiny±3.0}, 61.7{\tiny±2.3}) & ({\color[HTML]{FF0000} \textbf{70.0{\tiny±2.7}}}, {\color[HTML]{FF0000} \textbf{67.4{\tiny±1.8}}}) \\
                                  &                                      & \textbf{CIFAR10$_1$}        & (83.6{\tiny±0.6}, 97.3{\tiny±0.2})  & -                    & (63.9{\tiny±1.0}, 93.4{\tiny±0.3}) & (92.7{\tiny±0.1}, 99.0{\tiny±0.0})  & (78.1{\tiny±3.0}, 95.5{\tiny±0.1})   & (94.0{\tiny±1.0}, {\color[HTML]{0000FF} \textbf{99.2{\tiny±0.4}}}) & ({\color[HTML]{0000FF} \textbf{94.6{\tiny±0.6}}}, 99.0{\tiny±0.2}) & ({\color[HTML]{FF0000} \textbf{97.8{\tiny±0.4}}}, {\color[HTML]{FF0000} \textbf{99.7{\tiny±0.1}}}) \\
                                  &                                      & \textbf{CIFAR10$_2$}           & (63.1{\tiny±6.3}, 48.7{\tiny±4.7})  & -                    & (73.9{\tiny±1.1}, 60.6{\tiny±1.0}) & (91.5{\tiny±0.3}, 88.2{\tiny±0.9})  & (85.3{\tiny±2.0}, 77.9{\tiny±1.9})   & (92.8{\tiny±0.9}, 90.1{\tiny±2.0}) & ({\color[HTML]{0000FF} \textbf{93.3{\tiny±1.6}}}, {\color[HTML]{0000FF} \textbf{90.9{\tiny±1.3}}}) & ({\color[HTML]{FF0000} \textbf{97.1{\tiny±2.4}}}, {\color[HTML]{FF0000} \textbf{93.9{\tiny±1.6}}}) \\ \bottomrule
\end{tabular}
}
\caption{Image-level (AUROC, AP) results on ten real-world AD datasets under various few-shot AD settings. Best results and the second-best results are respectively highlighted in {\color[HTML]{FF0000} \textbf{red}} and {\color[HTML]{0000FF} \textbf{blue}}.}
\label{image}
\end{table*}    

\subsection{Main Results}
\subsubsection{Image-level FSAD performance}
Table~\ref{image} presents the image-level FSAD results of InCTRLv2, compared to seven SotA methods across ten AD datasets. The results show that InCTRLv2 significantly outperforms the SotA models across almost all datasets. Specifically, VLM-based methods consistently outperform their non-VLM counterparts, primarily due to the superior generalization capabilities inherent in pre-trained VLMs. Among non-VLM approaches, PatchCore achieves better performance than SPADE and PaDiM, but all three struggle to generalize across domains beyond their training distributions. In contrast, WinCLIP demonstrates improved cross-domain robustness by leveraging CLIP’s strong visual-language alignment. ResAD further advances performance by incorporating a residual learning mechanism that captures discrepancies more effectively.

Our previous InCTRL framework also adopts a residual learning strategy, specifically through in-context residual learning, and delivers significant gains over WinCLIP. Importantly, InCTRL precedes ResAD and does not utilize pixel-level anomaly masks for supervision, yet still achieves stronger generalization on cross-domain datasets (i.e., medical and semantic AD datasets), highlighting the broad transferability of in-context representations. By using larger ViT backbone (ViT-L-14-336), InCTRL+ further enhance the image-level performance in some datasets. Building upon this foundation, InCTRLv2 introduces precise pixel-level supervision and a novel normality-aware signal, leading to substantial improvements in image-level few-shot anomaly detection.

On average, InCTRLv2 outperforms the best competing methods by up to 3.9\% in AUROC and 4.5\% in AP on industrial anomaly detection datasets, 3.7\% in AUROC and 0.8\% in AP on medical datasets, and 5.7\% in AUROC and 6.6\% in AP on semantic anomaly detection datasets. This demonstrates the effectiveness of InCTRLv2 in enriching residual-based anomaly detection with complementary signals from semantic-guided in-context learning and normality-guided correction.

\subsubsection{Pixel-level FSAD performance}
Table~\ref{pixel} presents the pixel-level FSAD results, comparing InCTRLv2 against six state-of-the-art methods across five AD datasets. Similar patterns to the image-level results are observed. Among the Non-VLM-based methods, PaDiM and Patchcore outperform SPADE by leveraging neighborhood-aware patch-level representations. VLM-based methods demonstrate stronger generalization and achieve higher performance, benefiting from the superior generalization capabilities learned from large-scale pre-training. WinCLIP performs comparably to ResAD in the 1-shot setting, but its performance gains plateau as the number of normal prompts increases, whereas ResAD continues to improve.

In contrast, InCTRLv2 consistently outperforms all baselines across nearly all datasets, due to its integration of semantic-guided in-context learning and normality-guided correction, which jointly provide complementary semantic and normality-only cues that refine residual-based representations and enhance pixel-level anomaly localization. 
On average, compared to the best-performing baselines, InCTRLv2 achieves up to 1.5\% AUROC and 0.7\% AP on industrial AD datasets, and 16.3\% AUROC and 1.1\% AP on medical AD datasets.

\begin{table*}[ht]
\centering
\resizebox{\textwidth}{!}{
\begin{tabular}{c|c|c|ccc|ccc}
\toprule
\multirow{2}{*}{\textbf{Setup}}  & \multirow{2}{*}{\textbf{Domain}}     & \multirow{2}{*}{\textbf{Dataset}} & \multicolumn{3}{c|}{\textbf{Non-VLM-based Methods}}                 & \multicolumn{3}{c}{\textbf{VLM-based Methods}}               \\
                                 &                                      &                                   & \textbf{SPADE}       & \textbf{PaDiM}       & \textbf{Patchcore}    & \textbf{WinCLIP}     & \textbf{ResAD} & \textbf{InCTRLv2}    \\ \midrule
\multirow{5}{*}{\textbf{1-shot}} & \multirow{4}{*}{\textbf{Industrial}} & \textbf{VisA}                     & (62.8{\scriptsize±1.2}, 28.7{\scriptsize±1.2}) & (87.5{\scriptsize±0.6}, 60.7{\scriptsize±1.1}) & ({\color[HTML]{0000FF} \textbf{95.0{\scriptsize±0.3}}}, 79.9{\scriptsize±0.3})  & (94.7{\scriptsize±0.1}, {\color[HTML]{0000FF} \textbf{80.3{\scriptsize±0.3}}}) & (90.1{\scriptsize±0.3}, 68.9{\scriptsize±0.4})   & ({\color[HTML]{FF0000} \textbf{95.3{\scriptsize±0.1}}}, {\color[HTML]{FF0000} \textbf{87.8{\scriptsize±0.2}}}) \\
                                 &                                      & \textbf{MVTecAD}                  & (61.7{\scriptsize±1.3}, 24.4{\scriptsize±0.9}) & (88.8{\scriptsize±0.2}, 70.3{\scriptsize±0.3}) & (92.7{\scriptsize±0.3}, 77.2{\scriptsize±0.3})  & ({\color[HTML]{0000FF} \textbf{93.7{\scriptsize±0.1}}}, {\color[HTML]{0000FF} \textbf{84.2{\scriptsize±0.2}}}) & (90.6{\scriptsize±1.2}, 81.8{\scriptsize±1.4})   & ({\color[HTML]{FF0000} \textbf{95.2{\scriptsize±0.1}}}, {\color[HTML]{FF0000} \textbf{90.4{\scriptsize±0.2}}}) \\
                                 &                                      & \textbf{SDD}                      & (55.1{\scriptsize±0.3}, 19.9{\scriptsize±0.2}) & (89.9{\scriptsize±1.3}, 67.5{\scriptsize±3.6}) & (95.3{\scriptsize±0.7}, 82.5{\scriptsize±1.2})  & ({\color[HTML]{0000FF} \textbf{97.2{\scriptsize±0.7}}}, {\color[HTML]{0000FF} \textbf{85.3{\scriptsize±0.7}}}) & ({\color[HTML]{0000FF} \textbf{97.2{\scriptsize±0.0}}}, 83.9{\scriptsize±1.3})   & ({\color[HTML]{FF0000} \textbf{97.4{\scriptsize±0.1}}}, {\color[HTML]{FF0000} \textbf{94.5{\scriptsize±0.6}}}) \\
                                 &                                      & \textbf{AITEX}                    & (55.7{\scriptsize±3.8}, 31.8{\scriptsize±1.1}) & (71.4{\scriptsize±9.7}, 63.2{\scriptsize±6.2}) & (68.9{\scriptsize±1.2}, 31.4{\scriptsize±19.1}) & (74.1{\scriptsize±3.6}, 53.2{\scriptsize±4.5}) & ({\color[HTML]{FF0000} \textbf{88.9{\scriptsize±1.5}}}, {\color[HTML]{FF0000} \textbf{78.7{\scriptsize±2.3}}})   & ({\color[HTML]{0000FF} \textbf{83.5{\scriptsize±1.5}}}, {\color[HTML]{0000FF} \textbf{71.2{\scriptsize±1.7}}}) \\ \cmidrule{2-9} 
                                 & \textbf{Medical}                     & \textbf{BraTS}                    & (72.4{\scriptsize±1.4}, 29.7{\scriptsize±1.2}) & (91.7{\scriptsize±0.2}, 69.5{\scriptsize±0.4}) & ({\color[HTML]{0000FF} \textbf{96.4{\scriptsize±0.9}}}, {\color[HTML]{0000FF} \textbf{80.8{\scriptsize±3.5}}})  & (93.5{\scriptsize±0.3}, 71.0{\scriptsize±0.5}) & (92.2{\scriptsize±1.3}, 68.5{\scriptsize±4.6})   & ({\color[HTML]{FF0000} \textbf{97.1{\scriptsize±0.1}}}, {\color[HTML]{FF0000} \textbf{81.6{\scriptsize±0.5}}}) \\ \midrule
\multirow{5}{*}{\textbf{2-shot}} & \multirow{4}{*}{\textbf{Industrial}} & \textbf{VisA}                     & (62.8{\scriptsize±1.2}, 28.7{\scriptsize±1.2}) & (90.9{\scriptsize±0.5}, 69.4{\scriptsize±1.5}) & ({\color[HTML]{0000FF} \textbf{95.5{\scriptsize±0.2}}}, {\color[HTML]{0000FF} \textbf{82.5{\scriptsize±0.5}}})  & (95.0{\scriptsize±0.1}, 80.9{\scriptsize±0.3}) & (96.1{\scriptsize±0.9}, 81.3{\scriptsize±3.7})   & ({\color[HTML]{FF0000} \textbf{96.4{\scriptsize±0.0}}}, {\color[HTML]{FF0000} \textbf{89.7{\scriptsize±0.1}}}) \\
                                 &                                      & \textbf{MVTecAD}                  & (61.7{\scriptsize±1.3}, 24.4{\scriptsize±0.9}) & (91.4{\scriptsize±0.8}, 77.0{\scriptsize±1.6}) & (93.5{\scriptsize±0.2}, 79.8{\scriptsize±0.2})  & (93.9{\scriptsize±0.1}, 84.7{\scriptsize±0.2}) & ({\color[HTML]{0000FF} \textbf{94.0{\scriptsize±1.7}}}, {\color[HTML]{0000FF} \textbf{86.3{\scriptsize±2.1}}})   & ({\color[HTML]{FF0000} \textbf{95.7{\scriptsize±0.1}}}, {\color[HTML]{FF0000} \textbf{91.0{\scriptsize±0.1}}}) \\
                                 &                                      & \textbf{SDD}                      & (55.1{\scriptsize±0.3}, 19.9{\scriptsize±0.2}) & (92.9{\scriptsize±0.9}, 75.3{\scriptsize±2.0}) & (95.1{\scriptsize±0.3}, 82.9{\scriptsize±0.7})  & ({\color[HTML]{0000FF} \textbf{97.4{\scriptsize±0.5}}}, {\color[HTML]{0000FF} \textbf{85.4{\scriptsize±0.8}}}) & (97.2{\scriptsize±0.3}, 83.2{\scriptsize±0.8})   & ({\color[HTML]{FF0000} \textbf{97.6{\scriptsize±0.1}}}, {\color[HTML]{FF0000} \textbf{94.6{\scriptsize±0.2}}}) \\
                                 &                                      & \textbf{AITEX}                    & (59.7{\scriptsize±2.7}, 32.8{\scriptsize±4.9}) & (82.0{\scriptsize±2.2}, 74.5{\scriptsize±2.6}) & (77.7{\scriptsize±1.9}, 43.0{\scriptsize±3.6})  & (79.7{\scriptsize±1.3}, 55.8{\scriptsize±2.8}) & ({\color[HTML]{FF0000} \textbf{90.0{\scriptsize±2.3}}}, {\color[HTML]{FF0000} \textbf{75.5{\scriptsize±1.0}}})   & ({\color[HTML]{0000FF} \textbf{87.5{\scriptsize±1.3}}}, {\color[HTML]{0000FF} \textbf{73.4{\scriptsize±0.9}}}) \\ \cmidrule{2-9} 
                                 & \textbf{Medical}                     & \textbf{BraTS}                    & (72.4{\scriptsize±1.4}, 29.8{\scriptsize±1.2}) & (92.9{\scriptsize±0.5}, 73.5{\scriptsize±1.4}) & ({\color[HTML]{0000FF} \textbf{96.7{\scriptsize±0.3}}}, {\color[HTML]{0000FF} \textbf{82.8{\scriptsize±0.6}}})  & (94.2{\scriptsize±0.6}, 74.1{\scriptsize±1.1}) & (93.4{\scriptsize±0.6}, 69.0{\scriptsize±1.2})   & ({\color[HTML]{FF0000} \textbf{97.3{\scriptsize±0.0}}}, {\color[HTML]{FF0000} \textbf{83.9{\scriptsize±0.2}}}) \\ \midrule
\multirow{5}{*}{\textbf{4-shot}} & \multirow{4}{*}{\textbf{Industrial}} & \textbf{VisA}                     & (62.8{\scriptsize±1.2}, 28.7{\scriptsize±1.2}) & (92.2{\scriptsize±0.2}, 73.0{\scriptsize±0.4}) & (95.8{\scriptsize±0.3}, {\color[HTML]{0000FF} \textbf{83.1{\scriptsize±0.5}}})  & (95.1{\scriptsize±0.1}, 81.1{\scriptsize±0.3}) & ({\color[HTML]{0000FF} \textbf{96.7{\scriptsize±1.3}}}, {\color[HTML]{0000FF} \textbf{83.1{\scriptsize±5.0}}})   & ({\color[HTML]{FF0000} \textbf{96.7{\scriptsize±0.0}}}, {\color[HTML]{FF0000} \textbf{90.4{\scriptsize±0.1}}}) \\
                                 &                                      & \textbf{MVTecAD}                  & (61.7{\scriptsize±1.3}, 25.2{\scriptsize±1.9}) & (91.6{\scriptsize±0.9}, 77.4{\scriptsize±1.6}) & (94.5{\scriptsize±0.3}, 82.1{\scriptsize±0.8})  & (94.3{\scriptsize±0.1}, 85.3{\scriptsize±0.4}) & ({\color[HTML]{0000FF} \textbf{94.6{\scriptsize±2.1}}}, {\color[HTML]{0000FF} \textbf{88.4{\scriptsize±2.7}}})   & ({\color[HTML]{FF0000} \textbf{96.0{\scriptsize±0.1}}}, {\color[HTML]{FF0000} \textbf{91.7{\scriptsize±0.3}}}) \\
                                 &                                      & \textbf{SDD}                      & (55.1{\scriptsize±0.3}, 19.9{\scriptsize±0.2}) & (93.8{\scriptsize±0.3}, 78.0{\scriptsize±0.4}) & (95.5{\scriptsize±0.5}, 83.2{\scriptsize±1.6})  & ({\color[HTML]{0000FF} \textbf{98.1{\scriptsize±0.4}}}, {\color[HTML]{0000FF} \textbf{85.7{\scriptsize±0.6}}}) & (97.3{\scriptsize±0.5}, 83.2{\scriptsize±0.7})   & ({\color[HTML]{FF0000} \textbf{98.2{\scriptsize±0.1}}}, {\color[HTML]{FF0000} \textbf{95.0{\scriptsize±0.1}}}) \\
                                 &                                      & \textbf{AITEX}                    & (54.8{\scriptsize±1.6}, 39.1{\scriptsize±2.8}) & (82.7{\scriptsize±1.3}, 79.7{\scriptsize±3.1}) & (83.2{\scriptsize±1.6}, 50.9{\scriptsize±0.4})  & (82.2{\scriptsize±0.9}, 61.0{\scriptsize±1.1}) & ({\color[HTML]{FF0000} \textbf{92.7{\scriptsize±0.3}}}, {\color[HTML]{FF0000} \textbf{80.0{\scriptsize±1.8}}})   & ({\color[HTML]{0000FF} \textbf{91.4{\scriptsize±0.3}}}, {\color[HTML]{0000FF} \textbf{80.3{\scriptsize±0.7}}}) \\ \cmidrule{2-9} 
                                 & \textbf{Medical}                     & \textbf{BraTS}                    & (72.4{\scriptsize±1.4}, 29.7{\scriptsize±1.2}) & (94.5{\scriptsize±0.2}, 78.6{\scriptsize±0.6}) & ({\color[HTML]{0000FF} \textbf{96.9{\scriptsize±0.3}}}, {\color[HTML]{0000FF} \textbf{83.0{\scriptsize±1.1}}})  & (94.5{\scriptsize±0.4}, 74.5{\scriptsize±0.3}) & (93.0{\scriptsize±0.0}, 66.9{\scriptsize±0.7})   & ({\color[HTML]{FF0000} \textbf{97.5{\scriptsize±0.1}}}, {\color[HTML]{FF0000} \textbf{84.0{\scriptsize±0.1}}}) \\ \bottomrule
\end{tabular}}
\caption{Pixel-level (AUROC, PRO) results on five real-world AD datasets under various few-shot AD settings. Best results and the second-best results are respectively highlighted in {\color[HTML]{FF0000} \textbf{red}} and {\color[HTML]{0000FF} \textbf{blue}}.}
\label{pixel}
\end{table*}

\subsection{Ablation Study}
In our original work, we validated the importance of three core components in InCTRL, including text–prompt–guided features, patch-level residuals, and image-level residuals for enhancing generalization. To further assess the impact of the additional optimization terms introduced in InCTRLv2, we conduct an ablation study of their combinations in the four-shot setting. Since the OASL module operates in parallel with DASL, we first compare performance with and without the $\mathcal{L}_{OASL}$ term to assess its effect. We then investigate the contributions of the two optimization objectives ($\mathcal{L}_{P}$ and $\mathcal{L}_{I}$) within the DASL module.

\subsubsection{Image-level Ablation Study}
The image-level results of ablation study are reported in Table~\ref{ab_study}. Since image-level anomaly scores are generated solely within the DASL module, the inclusion or exclusion of $\mathcal{L}_{OASL}$ does not directly affect image-level prediction performance. Specifically, both $\mathcal{L}_{P}$ and $\mathcal{L}_{I}$ contribute to image-level anomaly detection, with $\mathcal{L}_{I}$ providing the dominant improvement by enforcing holistic anomaly discrimination that is essential for reliable classification. Meanwhile, $\mathcal{L}_{P}$ offers additional gains by injecting fine-grained anomaly cues that complement the global features. When optimized together, these two objectives lead to more balanced and robust image-level performance, demonstrating their complementary roles in enhancing the effectiveness of InCTRLv2.

\begin{table*}[t]
\centering
\resizebox{\textwidth}{!}{
\begin{tabular}{c|cc|cccc|ccc|cccc}
\toprule
\multirow{3}{*}{$\mathcal{L}_{OASL}$} & \multirow{3}{*}{$\mathcal{L}_{P}$} & \multirow{3}{*}{$\mathcal{L}_{I}$} & \multicolumn{4}{c|}{\textbf{Industrial}}                                                                                          & \multicolumn{3}{c|}{\textbf{Medical}}                                                                   & \multicolumn{4}{c}{\textbf{Semantical}}                                                                                            \\ \cmidrule{4-14} 
                              &                                    &                                     & \multirow{2}{*}{\textbf{VisA}} & \multirow{2}{*}{\textbf{SDD}} & \multirow{2}{*}{\textbf{ELPV}} & \multirow{2}{*}{\textbf{AITEX}} & \multirow{2}{*}{\textbf{BrainMRI}} & \multirow{2}{*}{\textbf{HeadCT}} & \multirow{2}{*}{\textbf{BraTS}} & \multicolumn{2}{c|}{\textbf{MINIST}}                             & \multicolumn{2}{c}{\textbf{CIFAR-10}}                            \\ \cmidrule{11-14} 
                              &                                  &                                     &                                &                               &                              &                                 &                                    &                                  &                                 & \textbf{1 vs All}    & \multicolumn{1}{c|}{\textbf{Multi-class}} & \textbf{1 vs All}    & \multicolumn{1}{c}{\textbf{Multi-class}} \\ \midrule
\multirow{4}{*}{NA}   & $\times$                         & $\times$                          & (89.1, 90.9)                   & (96.6, 94.0)                  & (73.2, 84.9)                   & (70.0, 49.5)                    & (97.8, {\color[HTML]{0000FF} \textbf{99.6}})                       & (88.0, 97.1)                     & (77.6, 96.0)                    & (83.6, 87.9) &   (60.3, 58.4)                                        &  (87.8, 93.4) &    (88.5, 89.0)                                      \\ \cmidrule{2-3}
   & \checkmark                         & $\times$                          & (90.8, 91.2)                   & (98.4, {\color[HTML]{0000FF} \textbf{96.1}})                  & (83.9, 91.4)                   & ({\color[HTML]{0000FF} \textbf{80.8}}, {\color[HTML]{0000FF} \textbf{63.0}})                    & (97.5, {\color[HTML]{0000FF} \textbf{99.6}})                       & (89.3, 97.2)                     & (86.1, 97.3)                    & (87.9, 96.2)         & \multicolumn{1}{c}{({\color[HTML]{0000FF} \textbf{68.3}}, 65.8)}          & (94.6, 98.3)         & \multicolumn{1}{c}{({\color[HTML]{0000FF} \textbf{95.1}}, 93.2)}         \\ \cmidrule{2-3}
                              & $\times$                         & \checkmark                          & ({\color[HTML]{0000FF} \textbf{91.2}}, {\color[HTML]{0000FF} \textbf{92.5}})                   & ({\color[HTML]{0000FF} \textbf{98.5}}, 95.9)                  & ({\color[HTML]{0000FF} \textbf{84.0}}, {\color[HTML]{FF0000} \textbf{91.9}})                   & (79.2, 62.8)                    & (98.2, {\color[HTML]{FF0000} \textbf{99.7}})                       & ({\color[HTML]{0000FF} \textbf{93.4}}, {\color[HTML]{0000FF} \textbf{98.0}})                     & ({\color[HTML]{0000FF} \textbf{87.9}}, {\color[HTML]{0000FF} \textbf{97.7}})                    & (88.4, {\color[HTML]{0000FF} \textbf{97.3}})         & \multicolumn{1}{c}{(67.9, {\color[HTML]{0000FF} \textbf{66.5}})}          & ({\color[HTML]{0000FF} \textbf{95.1}}, {\color[HTML]{0000FF} \textbf{98.4}})         & \multicolumn{1}{c}{(95.0, {\color[HTML]{0000FF} \textbf{93.3}})}         \\ \cmidrule{2-3}
                              & \checkmark                         & \checkmark                          & ({\color[HTML]{FF0000} \textbf{92.0}}, {\color[HTML]{FF0000} \textbf{93.1}})                   & ({\color[HTML]{FF0000} \textbf{98.9}}, {\color[HTML]{FF0000} \textbf{96.7}})                  & ({\color[HTML]{FF0000} \textbf{84.6}}, 91.6)                   & ({\color[HTML]{FF0000} \textbf{82.6}}, {\color[HTML]{FF0000} \textbf{63.4}})                    & ({\color[HTML]{FF0000} \textbf{98.4}}, {\color[HTML]{FF0000} \textbf{99.7}})                       & ({\color[HTML]{FF0000} \textbf{93.7}}, {\color[HTML]{FF0000} \textbf{98.4}})                     & ({\color[HTML]{FF0000} \textbf{89.5}}, {\color[HTML]{FF0000} \textbf{98.9}})                    & ({\color[HTML]{FF0000} \textbf{90.4}}, {\color[HTML]{FF0000} \textbf{98.3}})         & \multicolumn{1}{c}{({\color[HTML]{FF0000} \textbf{70.0}}, {\color[HTML]{FF0000} \textbf{67.4}})}          & ({\color[HTML]{FF0000} \textbf{97.8}}, {\color[HTML]{FF0000} \textbf{99.7}})         & \multicolumn{1}{c}{({\color[HTML]{FF0000} \textbf{97.1}}, {\color[HTML]{FF0000} \textbf{93.9}})}         \\ \bottomrule
\end{tabular}}
\caption{Image-level AUROC and AP results for ablation study under four-shot setting. Best results and the second-best results are respectively in {\color[HTML]{FF0000} \textbf{red}} and {\color[HTML]{0000FF} \textbf{blue}}. The results for VisA, and the one-vs-all settings of MNIST and CIFAR-10 represent an average result across their respective data subsets.}
\label{ab_study}
\end{table*}

\begin{table}[t]
\centering
\resizebox{0.48\textwidth}{!}{
\begin{tabular}{c|cc|ccc|c}
\toprule
\multirow{2}{*}{$\mathcal{L}_{OASL}$} & \multirow{2}{*}{$\mathcal{L}_{P}$} & \multirow{2}{*}{$\mathcal{L}_{I}$} & \multicolumn{3}{c|}{\textbf{Industrial}}       & \textbf{Medical} \\ \cmidrule{4-7}
                              &                                    &                                     & \textbf{VisA} & \textbf{SDD} & \textbf{AITEX} & \textbf{BraTS}   \\
              \midrule
\multirow{4}{*}{$\times$}            & $\times$                                  & $\times$                                   & (94.1, 86.8)                   & (95.8, 84.9)                  & (81.2, 71.3)                    & (96.3, 80.7)                    \\ \cmidrule{2-3}
                              & \checkmark                                  & $\times$                                   & (95.3, 89.4)                   & (97.4, 94.4)                  & (85.7, 74.4)                    & (96.8, 81.7)                    \\ \cmidrule{2-3}
                              & $\times$                                  & \checkmark                                   & (94.4, 86.2)                   & (95.9, 85.8)                  & (83.4, 71.5)                    & (96.2, 79.9)                    \\ \cmidrule{2-3}
                              & \checkmark                                  & \checkmark                                   & (95.7, 89.3)                   & (96.8, 89.0)                  & (87.6, 76.7)                    & (96.9, 82.1)                    \\   \midrule
\multirow{4}{*}{\checkmark}    & $\times$                                  & $\times$           & (94.8, 87.1)  & (97.5, 92.5) & (88.0, 76.6)   & (97.1, 82.0) \\ \cmidrule{2-3}      

& \checkmark                                  & $\times$                                   & ({\color[HTML]{0000FF} \textbf{96.2}}, {\color[HTML]{0000FF} \textbf{89.7}})                   & ({\color[HTML]{0000FF} \textbf{97.7}}, {\color[HTML]{0000FF} \textbf{94.5}})                  & ({\color[HTML]{0000FF} \textbf{89.0}}, {\color[HTML]{0000FF} \textbf{77.4}})                    & ({\color[HTML]{0000FF} \textbf{97.3}}, {\color[HTML]{0000FF} \textbf{82.3}})                    \\ \cmidrule{2-3}
                              & $\times$                                  & \checkmark                                   & (95.5, 88.9)                   & (97.6, 92.6)                  & (86.7, 76.6)                    & (97.1, {\color[HTML]{0000FF} \textbf{82.3}})                    \\  \cmidrule{2-3}
                              & \checkmark                                  & \checkmark                                   & ({\color[HTML]{FF0000} \textbf{96.7}}, {\color[HTML]{FF0000} \textbf{90.4}})                   & ({\color[HTML]{FF0000} \textbf{98.2}}, {\color[HTML]{FF0000} \textbf{95.0}})                  & ({\color[HTML]{FF0000} \textbf{91.4}}, {\color[HTML]{FF0000} \textbf{80.3}})                    & ({\color[HTML]{FF0000} \textbf{97.5}}, {\color[HTML]{FF0000} \textbf{84.0}})                    \\ \bottomrule
\end{tabular}}
\caption{Pixel-level AUROC and AP results for ablation study under four-shot setting. Best results and the second-best results are respectively in {\color[HTML]{FF0000} \textbf{red}} and {\color[HTML]{0000FF} \textbf{blue}}. The results for VisA represents an average result across their respective data subsets}
\label{ab_2}
\end{table}

\subsubsection{Pixel-level Ablation Study}
Since the pixel-level anomaly maps are determined by both the DASL and OASL modules, this ablation study investigates the effects of different combinations of the three optimization terms. The corresponding results are reported in Table~\ref{ab_2}. Specifically, incorporating $\mathcal{L}_{OASL}$ consistently improves performance over the variant without it, demonstrating the importance of the OASL module in enhancing the generalization ability of InCTRLv2. Moreover, $\mathcal{L}_{P}$ contributes more significantly than $\mathcal{L}_{I}$ to pixel-level performance, since $\mathcal{L}_{P}$ directly benefits from mask-level supervision to capture fine-grained anomaly cues, whereas the effect of $\mathcal{L}_{I}$ is relatively limited due to the absence of explicit pixel-level ground-truth guidance. The complete InCTRLv2 achieves the best results, demonstrating that the three optimization terms play complementary roles in improving the anomaly localization effectiveness of InCTRLv2.

\begin{figure}[ht]
    \centering
    \includegraphics[width=0.47\textwidth]{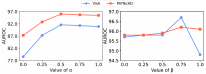}
    \caption{\textbf{Left}: Image-level AUROC results based on different value of $\alpha$. \textbf{Right}: Pixel-level AUROC results based on different value of $\beta$.}
    \label{fig:hyper}
\end{figure}

\begin{figure}[ht]
    \centering
    \includegraphics[width=0.45\textwidth]{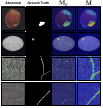}
    \caption{Visualization of anomaly maps generated by DASL module ($\mathbf{M}_p$) alone and InCTRLv2 ($\mathbf{M}$). The anomaly score maps are generated under the `VisA to MVTecAD' setting.}
    \label{fig:OASL}
\end{figure}

\subsection{Hyperparameter Sensitivity Analysis}
We analyze the effect of two key hyperparameters in InCTRLv2: $\alpha$ and $\beta$. $\alpha$ controls the contribution of the maximum patch-level residual score from $\mathbf{M}_x$ to the final image-level anomaly score (Eq.~\ref{eq:finalscore}), while $\beta$ regulates the weight of the normality-aware anomaly map produced by the OASL module in the final pixel-level anomaly map (Eq.~\ref{eq:final-map}). The corresponding results are shown in Fig. \ref{fig:hyper}.

Specifically, for $\alpha$, we observe that performance improves as the value increases, typically peaking at $\alpha = 0.5$. Beyond this point, the performance slightly declines, suggesting that incorporating patch-level residual information is crucial for robust image-level anomaly detection, but excessive reliance on patch-level cues can overshadow global semantics. In contrast, the pixel-level AUROC remains relatively stable across different values of $\beta$, suggesting that the OASL module consistently contributes a strong normality-aware signal regardless of its weighting. The best overall performance is obtained at $\beta = 0.75$, which achieves an effective balance between the anomaly map from the main branch and the normality-aware anomaly map from OASL.

\subsection{Qualitative Results}
\subsubsection{Visualization Study of OASL}
To further validate the correction contribution of the OASL module, we present qualitative comparisons between the anomaly map generated by DASL alone ($\mathbf{M}_p$, from Eq.~\ref{eq:3}) and the integrated anomaly map from DASL and OASL (InCTRLv2), denoted as $\mathbf{M}$ (Eq.~\ref{eq:final-map}). 

As shown in Fig.~\ref{fig:OASL}, $M_p$ often suffers from two major issues: (i) false positive responses on normal regions, as observed in the first two rows, and (ii) fragmented or incomplete localization of true anomalies, as shown in the last two rows. With the integration of the OASL module ($M$), the anomaly maps become more accurate and coherent: normal regions are better suppressed, while abnormal regions are highlighted with improved continuity and precision. This demonstrates that OASL effectively corrects semantic-guided predictions from DASL by introducing a stable normality-guided signal, thereby reducing false alarms and enhancing localization quality.

\subsubsection{Visualization Comparison with SotA models}
\begin{figure}[ht]
    \centering
    \includegraphics[width=0.45\textwidth]{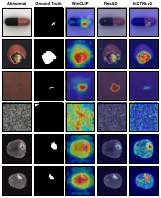}
    \caption{Visualization of anomaly maps generated by different GADS methods. The anomaly score maps are generated under the `VisA to MVTecAD' setting.}
    \label{fig:qr}
\end{figure}

We also compare the pixel-level anomaly maps generated by InCTRLv2 with those produced by other VLM-based FSAD models across multiple datasets, as shown in Fig.~\ref{fig:qr}. InCTRLv2 achieves substantially more accurate anomaly localization and markedly reduces false positives in normal regions, outperforming competing methods across both object and textual defective data. Notably, despite not leveraging any additional supervision or training on medical data, InCTRLv2 effectively localizes abnormal lesion and tumor regions. This highlights the strong cross-dataset generalization achieved by InCTRLv2 through its transferable in-context knowledge.

\begin{table}[t]
\centering
\resizebox{0.4\textwidth}{!}{
\begin{tabular}{c|cc}
\hline
      \multicolumn{1}{c|}{{\color[HTML]{333333} \textbf{Methods}}}                                    & \multicolumn{1}{c}{{\color[HTML]{333333} \textbf{Parameters(M)}}} & \multicolumn{1}{c}{{\color[HTML]{333333} \textbf{Infer time(fps)}}} \\ \hline
{\color[HTML]{333333} \textbf{SPADE}}     & {\color[HTML]{333333} 74.5}                                       & {\color[HTML]{333333} 4.8}                                          \\
{\color[HTML]{333333} \textbf{PaDiM}}     & {\color[HTML]{333333} 686.9}                                      & {\color[HTML]{333333} 14.1}                                         \\
{\color[HTML]{333333} \textbf{PatchCore}} & {\color[HTML]{333333} 69.5}                                       & 21.5                                                                \\
{\color[HTML]{333333} \textbf{WinCLIP}}   & {\color[HTML]{333333} 165.9}                                      & 0.51                                                                \\
{\color[HTML]{333333} \textbf{ResAD}}     & 442.6                                                             & 18.8                                                                \\
{\color[HTML]{333333} \textbf{InCTRL}}    & 117.5                                                             & 0.53                                                                \\
\textbf{InCTRLv2}                       & 437.6                                                               & 5.86                                                                \\ \hline
\end{tabular}}
\caption{Complexity comparison between our InCTRLv2 and other competing methods}
\label{time_complexity}
\end{table}

\subsection{Model Complexity Comparison}
We compare the model complexity of InCTRLv2 with several state-of-the-art methods in Table~\ref{time_complexity}, evaluating both the number of parameters and the per-image inference time. The parameter count of InCTRLv2 is comparable to other CLIP-based approaches, indicating that our method does not rely on an excessively large model capacity. Since training is conducted entirely offline, the additional computational overhead during training is generally negligible in practical deployments. At inference, InCTRLv2 maintains reasonable efficiency and responsiveness, achieving a favorable trade-off between accuracy and computational cost.

\section{Conclusion}
In this work, we propose InCTRLv2, an enhanced generalist anomaly detection framework that extends our prior in-context residual learning model 
with a dual-branch framework featuring two key innovations: a discriminative anomaly score learning module for robust semantic-guided anomaly discrimination, and a one-class anomaly score learning module to extract domain-invariant normality patterns.
Extensive experiments across ten diverse anomaly detection benchmarks and multiple few-shot settings show that InCTRLv2 achieves state-of-the-art performance in both image-level detection and pixel-level localization. These results underscore the generalization capability, interpretability, and robustness of the proposed design, making it a strong candidate for practical deployment in open-world anomaly detection scenarios.

\section*{Acknowledgment}
This research is partially supported by A*STAR under its MTC YIRG Grant (M24N8c0103), the Singapore Ministry of Education (MOE) Academic Research Fund (AcRF) Tier 1 Grant (24-SIS-SMU-008), and the Lee Kong Chian Fellowship (T050273).

\backmatter

\bibliography{sn-ijcv-arxiv}

\end{document}